\documentclass[acmlarge]{acmart}

\makeatletter
\newcommand{\myconfshort}{\acmConference@shortname}
\newcommand{\myconffull}{\acmConference@name}
\newcommand{\myconfdate}{\acmConference@date}
\newcommand{\myconfloc}{\acmConference@venue}

\AtBeginDocument{%
  \fancypagestyle{firstpagestyle}{%
    \fancyhead{}%
    \fancyfoot[C]{}%
  }%
  \fancyhf{}%
  \fancyhead[LO]{\@headfootfont\shorttitle}%
  \fancyhead[RE]{\@headfootfont\@shortauthors}%
  \fancyhead[LE]{\@headfootfont\footnotesize \myconfshort, \myconfdate, \myconfloc}%
  \fancyhead[RO]{\@headfootfont\footnotesize \myconfshort, \myconfdate, \myconfloc}%
  \fancyfoot[C]{}%
}
\makeatother

\copyrightyear{2026}
\acmYear{2026}
\setcopyright{cc}
\setcctype{by}
\acmConference[FAccT '26]{The 2026 ACM Conference on Fairness, Accountability, and Transparency}{June 25--28, 2026}{Montreal, QC, Canada}
\acmBooktitle{The 2026 ACM Conference on Fairness, Accountability, and Transparency (FAccT '26), June 25--28, 2026, Montreal, QC, Canada}
\acmDOI{10.1145/3805689.3812229}
\acmISBN{979-8-4007-2596-8/2026/06}

\usepackage{nicefrac}
\usepackage{microtype}
\usepackage{subcaption}
\usepackage{adjustbox}
\usepackage{multirow}

\title[Energy Scaling Laws for Diffusion Models]
{Energy Scaling Laws for Diffusion Models: Quantifying Compute in Image Generation}

\author{Aniketh Iyengar}
\affiliation{%
  \institution{Stanford University}
  \city{Stanford}
  \country{USA}}
\email{aniketh@stanford.edu}

\author{Jiaqi Han}
\affiliation{%
  \institution{Stanford University}
  \city{Stanford}
  \country{USA}}
\email{jiaqihan@stanford.edu}

\author{Boris Ruf}
\affiliation{%
  \institution{AXA AI Research}
  \city{Paris}
  \country{France}}
\email{boris.ruf@axa.com}

\author{Vincent Grari}
\affiliation{%
  \institution{AXA AI Research}
  \city{Paris}
  \country{France}}
\email{vincent.grari@axa.com}

\author{Marcin Detyniecki}
\affiliation{%
  \institution{AXA AI Research}
  \city{Paris}
  \country{France}}
\email{marcin.detyniecki@axa.com}

\author{Stefano Ermon}
\affiliation{%
  \institution{Stanford University}
  \city{Stanford}
  \country{USA}}
\email{ermon@cs.stanford.edu}

\renewcommand{\shortauthors}{Iyengar et al.}

\begin{document}

\begin{abstract}
The rapidly growing computational demands of diffusion models for image generation have raised significant concerns about energy consumption and environmental impact. While existing approaches to energy optimization focus on architectural improvements or hardware acceleration, there is a lack of principled methods to predict energy consumption across different model configurations and hardware setups. We propose an adaptation of Kaplan scaling laws to predict GPU energy consumption for diffusion models based on computational complexity (FLOPs). Our approach decomposes diffusion model inference into text encoding, iterative denoising, and decoding components, with the hypothesis that denoising operations dominate energy consumption due to their repeated execution across multiple inference steps. We conduct comprehensive experiments across four state-of-the-art diffusion models (Stable Diffusion 2, Stable Diffusion 3.5, Flux, and Qwen) on three GPU architectures (NVIDIA A100, A4000, A6000), spanning various inference configurations including resolution ($256^2$-$1024^2$), precision ($\text{fp16}$/$\text{fp32}$), step counts ($10$-$50$), and classifier-free guidance settings. Our energy scaling law achieves high predictive accuracy within individual architectures ($R^2 > 0.9$) and exhibits strong cross-architecture generalization, maintaining high rank correlations across models and enabling reliable energy estimation for unseen model–hardware combinations. These results validate the compute-bound nature of diffusion inference and establish energy consumption estimation as a necessary foundation for sustainable AI deployment planning and subsequent carbon footprint assessment.
\end{abstract}




\maketitle

\section{Introduction}
\label{sec:introduction}

The rapid advancement of generative AI, particularly diffusion models~\cite{song2020score,ho2020denoising,sohl2015deep} for high-quality image synthesis, has transformed creative applications and scientific visualization. Models like Stable Diffusion \cite{rombach2022high}, DALL-E~\cite{ramesh2021zero}, and recent innovations such as Flux \cite{esser2024scalingrectifiedflowtransformers} and Gemini 2.5 Flash Image "Nano Banana" \cite{google2025gemini} can generate photorealistic images from text prompts with unprecedented quality. However, this capability comes at substantial computational cost—generating a single high-resolution image can require billions of floating-point operations across dozens of denoising steps.

As these models are deployed at scale, their energy consumption has become a critical concern. Recent studies estimate that training large language models can consume as much electricity as hundreds of homes over several months \cite{strubell2019energy}, and inference costs can be equally significant when scaled to billions of users. For diffusion models, the energy challenge is compounded by their iterative nature—unlike single forward-pass models, diffusion models require multiple denoising steps, each involving full network evaluation~\cite{ho2020denoising,song2020denoising}.

Despite the critical importance of energy efficiency, current approaches to estimating diffusion model energy consumption often rely on post-hoc empirical measurements or \textcolor{black}{specialized benchmarks \cite{seyfarth2025latent, bertazzini2025hidden}}. While these provide valuable ground-truth data, there remains a need for predictive models that relate architectural complexity to energy usage. The research landscape presents several challenges: a relative scarcity of energy modeling for diffusion models compared to language models, limited generalization across diverse model-hardware combinations, and the absence of theoretical frameworks that link FLOP counts directly to energy footprints across varying hardware configurations.

This work addresses these gaps by developing a principled approach to predicting diffusion model energy consumption based on scaling laws. Our key contributions are: (1) adaptation of Kaplan scaling laws \cite{kaplan2020scaling} to formulate energy scaling relationships for diffusion models, (2) comprehensive FLOP decomposition for four major diffusion architectures, (3) cross-architecture validation demonstrating generalization across different model families, (4) hardware-aware modeling incorporating precision and GPU architecture effects, and (5) a framework for estimating the energy impact of large-scale diffusion model deployment\footnote{\textcolor{black}{We release our codebase for reproducibility at \url{https://github.com/ani11452/energy_scaling_laws}}}. 

\textcolor{black}{Our empirical analysis reveals a near-linear relationship between computational complexity and GPU energy consumption across diffusion models, indicating that inference operates in a largely compute-bound regime. We further observe that these scaling relationships remain consistent across multiple model architectures and GPU platforms, enabling reliable energy estimation for unseen configurations. These findings suggest that FLOP-based models can provide practical guidance for deployment planning and energy-aware optimization of diffusion systems.} 

\textcolor{black}{The remainder of this paper is organized as follows. Section 2 covers related work and background, Section 3 presents the theoretical foundations, Section 4 details the methodology and experimental design, and Section 5 describes the experimental setup. We conclude with results and discussion.}

\section{Related Work}
\label{sec:related_work}

\textbf{Energy Efficiency in Deep Learning.} The environmental impact of machine learning has gained increasing attention. Strubell et al. \cite{strubell2019energy} demonstrated that training large transformer models can produce as much CO$_2$ as several cars over their lifetimes. Henderson et al. \cite{henderson2020towards} established frameworks for systematic carbon footprint reporting. Tools like CarbonTracker \cite{anthony2020carbontracker} and CodeCarbon \cite{codecarbon} provide energy monitoring, though they focus on measurement rather than prediction. Recent work by Luccioni et al. \cite{luccioni2022estimating, luccioni2023power} analyzed energy consumption for large models but remains reactive rather than predictive. Furthermore, a complete environmental assessment requires considering the full lifecycle of hardware and shared platforms \cite{guennebaud2025evaluating, morand2024mlca, berthelot2025understanding}. It is also critical to acknowledge potential rebound effects, where efficiency gains may paradoxically increase total energy consumption through increased usage \cite{luccioni2025efficiency}.

\textbf{Scaling Laws for Neural Networks.} Kaplan et al. \cite{kaplan2020scaling} established foundational scaling laws for transformer-based language models, demonstrating that performance scales predictably with parameters, dataset size, and compute budget. \textcolor{black}{The Chinchilla model introduced in \cite{hoffmann2022training}} refined these laws for optimal resource allocation. However, existing scaling laws focus on model \textcolor{black}{training and validation loss} rather than inference energy consumption, and primarily address autoregressive language models rather than diffusion-based generative models. \textcolor{black}{Recent work has begun to examine the environmental footprint of model inference more broadly—for example, Jegham et al.~\cite{jegham2025hungry} benchmark the energy, water, and carbon footprint of LLM inference workloads—but these studies focus on measurement rather than predictive modeling, and do not address the distinct computational structure of diffusion-based generation}.

\textbf{Diffusion Model Efficiency.} Computational efficiency research has largely focused on reducing latency rather than energy consumption. DDIM \cite{ho2020denoising} introduced deterministic sampling to reduce denoising steps, classifier-free guidance \cite{ho2022classifier} improved output quality at the cost of doubled computational requirements, and latent diffusion models \cite{rombach2022high} reduce cost by operating in compressed latent spaces. While early energy analyses relied on FLOP counting, recent work has introduced direct measurement techniques \cite{seyfarth2025latent, bertazzini2025hidden}. \textcolor{black}{Most relevant to this study, Delavande et al.~\cite{delavande2025videokilledenergybudget} provide an operator-level analysis of the computational and energy characteristics of diffusion models on H100 GPUs, offering fine-grained characterization of energy usage within specific model components and identifying quadratic scaling of spatio-temporal attention. Complementarily, this work pursues model-level predictive frameworks that estimate total inference energy from architectural metadata—such as FLOPs and deployment configuration—enabling energy prediction for unseen model–hardware combinations without requiring per-operator profiling or direct access to model implementations.} Lastly, progressive distillation \cite{salimans2022progressive, meng2023distillation} and few-step samplers \cite{lu2022dpm, lu2025dpmp, zhang2022gddim} represent promising efficiency directions, though energy benefits vary across hardware platforms.

\textbf{Hardware-Aware Machine Learning.} GPU architecture significantly impacts energy efficiency. Mixed-precision training \cite{micikevicius2017mixed} can reduce energy consumption while maintaining quality. Tensor core acceleration provides efficiency improvements that depend on workload characteristics \cite{markidis2018nvidia, jia2019dissecting}. Our scaling law approach provides a framework for estimating energy consumption across different hardware platforms by leveraging architectural trends, which can help inform resource allocation decisions.

\textbf{Diffusion Model Architecture.}
Diffusion models generate images through a learned denoising process that iteratively refines random noise into coherent images. Modern text-to-image diffusion models typically consist of three main components: text encoding, iterative denoising, and image decoding.
\begin{itemize}
    \item  \textbf{Text Encoding.} Input text prompts are processed by transformer-based encoders (e.g., CLIP \cite{radford2021learningtransferablevisualmodels}, T5 \cite{raffel2023exploringlimitstransferlearning}) to produce contextual embeddings that guide the generation process. This step occurs once per inference and typically represents a small fraction of total computation. 

    \item \textbf{Iterative Denoising.} The core computational process involves repeatedly applying a neural network to progressively denoise a latent representation. While many methods employ a U-Net \cite{ronneberger2015unetconvolutionalnetworksbiomedical} or standard Transformer backbone \cite{vaswani2023attentionneed}, state-of-the-art diffusion models typically use variants of the MMDiT architecture \cite{esser2024scalingrectifiedflowtransformers}. Each denoising iteration requires a full forward pass through the network, with typical inference comprising 10–50 such steps. When classifier-free guidance is applied, the computational cost effectively doubles—whether implemented as an expanded batch dimension or two sequential forward passes—since both conditional and unconditional evaluations are required \footnote{Some models implement amortized classifier-free guidance using a learned token. Here, we instead consider the true classifier-free guidance case from a computational footprint perspective.}. 

    \item \textbf{Image Decoding.} The final denoised latent representation is decoded into pixel space using a learned decoder network (e.g., variational autoencoder (VAE)). Like text encoding, this occurs once per inference.
\end{itemize}

\begin{figure}[ht]
    \centering
    \includegraphics[width=0.87\linewidth]{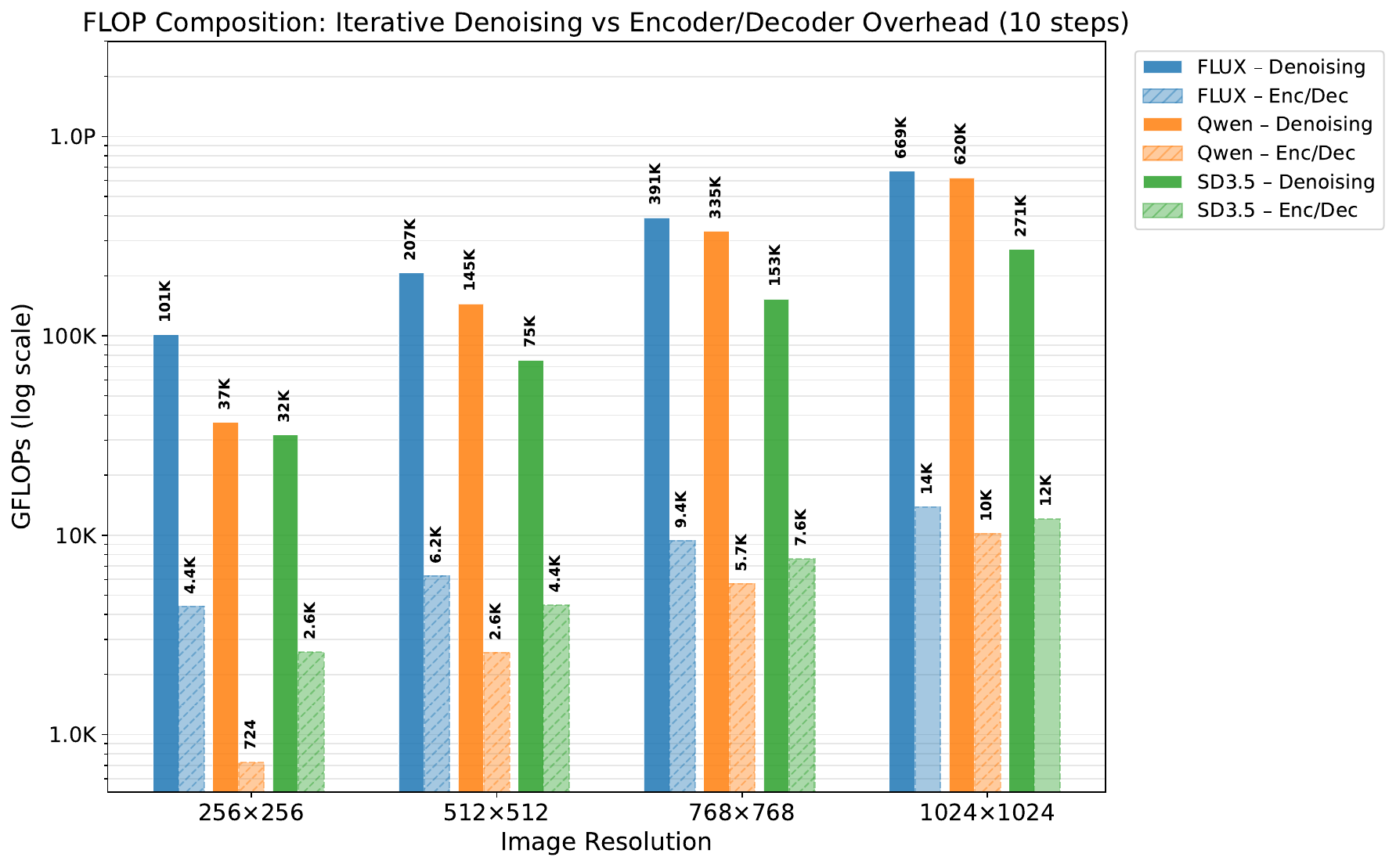}
    \caption{
    \textbf{Compute distribution across models and resolutions.}
  For each model (\textsc{Flux}, \textsc{Qwen}, and \textsc{SD\,3.5}), we plot the absolute GFLOPs attributed to the iterative denoising process (10 diffusion steps) versus the
  encoder/decoder overhead on a logarithmic scale.}
\label{fig:flop-decomp}
\end{figure}

The iterative nature of denoising makes it the dominant computational component, typically accounting for more than 90\% of inference FLOPs, as shown by our decomposition analysis in \textcolor{black}{Fig.~\ref{fig:flop-decomp}}.

\section{Methodology}
\label{sec:methodology}

\subsection{Kaplan Scaling Laws}
The original Kaplan scaling laws \cite{kaplan2020scaling} demonstrate that model performance scales predictably with computational resources according to power-law relationships: $L \propto P^{-\alpha_P}$, $L \propto C^{-\alpha_C}$, and $L \propto D^{-\alpha_D}$, where $L$ denotes loss, $P$ is the number of parameters, $C$ represents compute (in FLOPs), and $D$ is the dataset size. The scaling exponents $\alpha$ typically fall within the range $0.05$–$0.095$. These relationships provide a principled framework for resource allocation in large-scale language model development. 

Our work adapts this methodology to energy consumption, hypothesizing that energy usage follows power-law relationships with computational complexity: $E \propto \text{FLOPs}^{\alpha}$, albeit with different exponents and additional hardware-dependent factors. Kaplan et al.\ also introduced an analytical approximation for the FLOPs required per forward pass in transformer-based architectures, which we adopt in our analysis.

\subsection{GPU Energy Consumption}
GPU energy consumption depends on multiple factors including computational workload, memory access patterns, and hardware utilization. Modern GPUs are designed for high throughput parallel computation, with energy efficiency varying significantly based on workload characteristics.

\textbf{Computational vs. Memory Bound.} The energy efficiency of a deep learning workload is often determined by whether it is \textit{compute-bound} or \textit{memory-bound}~\cite{williams2009roofline}. Compute-bound operations, such as large matrix multiplications and convolutions, perform a high number of floating-point operations (FLOPs) for every byte of data transferred from memory. On modern GPUs, these operations can saturate the processor's arithmetic throughput, leading to higher energy efficiency as the hardware's fixed power overhead is amortized over more useful work. In contrast, memory-bound operations---such as element-wise transformations or data movement---are limited by memory bandwidth, often leaving the GPU's compute cores idle while waiting for data. Diffusion model inference is primarily compute-bound, driven by the intensive attention and convolution layers within its denoising backbone~\cite{esser2024scalingrectifiedflowtransformers}. Nevertheless, optimizing low-level kernel implementations to improve memory access and reduce unnecessary data transfers can further reduce energy consumption~\cite{dao2022flashattention, ivanov2021data}.

\textbf{Precision Effects.} Numerical precision significantly impacts both performance and energy. Transitioning from standard 32-bit floating-point ($\text{fp32}$) to lower-precision formats like 16-bit ($\text{fp16}$) or Brain Floating Point ($\text{bfloat16}$) reduces the amount of data moved across the memory bus by half. Furthermore, modern GPUs incorporate specialized \textit{Tensor Cores} dedicated to accelerating low-precision matrix operations, which provide substantial improvements in energy efficiency~\cite{micikevicius2017mixed, markidis2018nvidia, jia2019dissecting}.

\textbf{Hardware Architecture.} Different GPU architectures exhibit varying energy efficiency characteristics. Newer architectures generally provide better performance-per-watt ratios, but absolute energy consumption depends on workload size and utilization patterns.

Understanding these factors is crucial for developing accurate energy prediction models that generalize across different hardware platforms and computational workloads.

\subsection{FLOP Calculation for Diffusion Models}
For diffusion models, we decompose the total computational cost as:

\begin{equation}
\text{FLOPs}_{\text{total}} = \text{FLOPs}_{\text{text}} + T \times \text{FLOPs}_{\text{denoise}} + \text{FLOPs}_{\text{decode}}
\label{eq:flop_decomposition}
\end{equation}

This decomposition mirrors the structure of diffusion inference: text encoding and image decoding occur once per prompt, while denoising is repeated over $T$ iterations. Accordingly, we scale $\text{FLOPs}_{\text{denoise}}$ by $T$ to capture the dominant iterative cost. \textcolor{black}{We rely on analytical FLOP estimation rather than runtime FLOP counters, as many profiling tools require executing a forward pass and access to the model implementation. Our approach hopes to enables pre-deployment energy prediction directly from architectural metadata, which is particularly useful for closed-source models.}

Since the majority of FLOPs arise from denoising, the total compute for a denoising module with $P$ non-embedding parameters, attention dimension $d_{\text{attn}}$, $n_{\text{layers}}$ layers, and sequence length $L$ can be approximated using the Kaplan formulation~\cite{kaplan2020scaling}:

\begin{equation}
\text{FLOPs}_{\text{denoise}} \approx 2P + 2Ld_{\text{attn}}n_{\text{layers}}
\label{eq:transformer_flops}
\end{equation}

For diffusion models using U-Net architectures, FLOP calculation requires detailed analysis of convolution operations, attention mechanisms, and skip connections. We develop model-specific FLOP estimation functions that account for these architectural differences while maintaining the theoretical foundation established by Kaplan et al.~\citep{kaplan2020scaling}. As illustrated in Figure \ref{fig:flop-decomp}, even at a minimum of $T=10$ steps, the iterative denoising process accounts for over 90\% of the total computational budget across all models and resolutions. Since this proportion only increases as the step count $T$ grows, we can safely neglect the fixed overhead of text encoding and image decoding. \textbf{Consequently, for $N$ total queries, we approximate the total compute as $\text{FLOPs}_{\text{total}} \approx N \times T \times \text{FLOPs}_{\text{denoise}}$ in all subsequent experiments.}

\subsection{Energy Scaling Law Formulation}

We propose an energy scaling law inspired by Kaplan's approach, relating energy consumption $E$ to computational complexity (FLOPs) and key hardware and model configuration factors. In this work, we focus specifically on the dynamic energy consumption of the GPU, which represents the primary variable cost for large-scale model inference. Classifier-free guidance, floating-point precision, GPU architecture, and image resolution are incorporated as additive terms in a log-linear regression model, with FLOPs serving as the primary predictor of compute-bound energy usage. \textcolor{black}{This log-linear formulation corresponds to a multiplicative scaling law in the original (linear) space, where energy scales as a power law in FLOPs, with hardware and configuration factors acting as multiplicative modifiers.}

\begin{align}
\log(E) &= \log(A) + \alpha \log(\text{FLOPs} \times 2^{\mathbb{I}_{\text{cfg}}}) \nonumber\\
&\quad + \beta_{\text{dtype}} \mathbb{I}_{\text{dtype}} + \beta_{\text{gpu}} \mathbb{I}_{\text{gpu}} + \beta_{\text{res}} \log\left(\frac{H \times W}{256}\right)
\label{eq:energy_scaling_log}
\end{align}

\textcolor{black}{Learned values: $A$ is the constant within the power law, $\alpha$ is the power exponent of effective FLOPs, $\beta_{\text{dtype}}$ and $\beta_{\text{gpu}}$ quantify multiplicative shifts due to precision and hardware architecture, and $\beta_{\text{res}}$ accounts for image resolution bias.}

\subsection{Feature Engineering}
To implement our energy scaling law in practice, we construct a feature vector $\mathbf{x}$ that serves as the input to our linear regression model. Each component in $\mathbf{x}$ corresponds directly to a term in Equation~\ref{eq:energy_scaling_log}, enabling direct parameter estimation through ordinary least squares. Hardware and configuration indicators are encoded as one-hot variables, transforming categorical attributes (GPU type, precision) into numerical features suitable for regression. The feature vector consists of:

\begin{equation}
\mathbf{x} = [1, \log(\text{FLOPs}_{\text{cfg}}), \mathbb{I}_{\text{fp32}}, \mathbb{I}_{\text{A4000}}, \mathbb{I}_{\text{A6000}}, \log(H \times W / 256)]^T
\label{eq:feature_vector}
\end{equation}

Here, the intercept term (1) captures base energy consumption; $\text{FLOPs}_{\text{cfg}} = \text{FLOPs} \times 2^{\mathbb{I}_{\text{cfg}}}$ accounts for the doubling of denoising FLOPs when classifier-free guidance is enabled; $\mathbb{I}_{\text{fp32}}$ is a binary indicator for 32-bit precision (with 16-bit as the baseline); $\mathbb{I}_{\text{A4000}}$ and $\mathbb{I}_{\text{A6000}}$ are one-hot indicators for GPU architecture (with A100 as the baseline); and the resolution bias term, $\log(H \times W / 256)$, accounts for efficiency variations beyond pure FLOP scaling. In implementation, we define $\mathbb{I}_{\text{dtype}} := \mathbb{I}_{\text{fp32}}$ and $\mathbb{I}_{\text{gpu}} := [\mathbb{I}_{\text{A4000}}, \mathbb{I}_{\text{A6000}}]$ to match the theoretical formulation in Equation~\ref{eq:energy_scaling_log}. \textcolor{black}{Lastly, while one-hot encodings are suboptimal for generalizing to unseen inference configurations, we adopt them here given that standard GPU configurations are relatively limited in variety and continuous alternatives risk overfitting on the available data.}

The resolution bias term accounts for hardware-specific efficiency effects across tensor sizes. We hypothesize that energy efficiency varies with resolution due to differences in memory bandwidth utilization, cache behavior, and kernel-level optimizations not reflected in FLOP counts alone.

\section{Experimental Setup}
\label{sec:experimental_setup}

\subsection{Energy Measurement} We use CodeCarbon's EmissionsTracker~\cite{codecarbon} to monitor GPU power consumption at $1$Hz sampling rate throughout inference. CodeCarbon internally uses NVIDIA Management Library (NVML) to query GPU power consumption. Total energy is computed as the integral of power consumption over time, with baseline idle power subtracted to isolate inference-specific consumption. \textcolor{black}{As a note, estimating the energy consumption of AI workloads remains challenging in practice. Recent work has shown that commonly used tools such as CodeCarbon provide practical software-based estimates but can deviate from ground-truth measurements due to unobserved system-level factors such as cooling overheads and peripheral power consumption \cite{fischer2025groundtruthingaienergyconsumption}. These findings highlight the importance of interpreting energy estimates with awareness of their underlying assumptions. Nonetheless, we utilize it in our framework along with a batch size $=1$, following the methodology of \cite{Luccioni_2024}, reflecting deployment scenarios where requests are often processed sequentially rather than in large batches.}

\subsection{Validation Strategy} We employ two complementary validation approaches to assess generalization:

\textbf{Within-Architecture:} We apply 2-fold cross-validation on data from a single model-GPU combination (e.g., Flux on A100). This assesses scaling law stability when training and testing conditions are matched. Results are reported in Figure~\ref{fig:model_specific_validation_plots}.

\textbf{Cross-Architecture:} We train on one or more model-GPU combinations and test on held-out models or hardware platforms. For example, training on A100 data and testing on A6000 evaluates cross-GPU generalization (Figure~\ref{fig:across_gpus_validation_plots}), while training on Flux+SD3.5 data and testing on Qwen evaluates cross-model generalization (Figure~\ref{fig:cross_model_validation}). This strategy validates that our scaling laws capture architecture-agnostic energy-complexity relationships.

Each experimental configuration is run with a fixed random seed for reproducibility. We assess model performance using statistical measures including $R^2$, mean absolute error (MAE), and Spearman rank \& Pearson correlation coefficients to evaluate scaling law accuracy and generalization capability.

\subsection{Model Selection and FLOP Estimation}

We evaluate four representative diffusion models spanning distinct architectural families and compute FLOPs using model-specific formulas derived from architectural analysis (see Tables~\ref{tab:flop_formulas}, \ref{tab:model_flops_breakdown}, and \ref{tab:sd2_details}).

\textbf{Stable Diffusion 3.5-Large (SD3.5) \textcolor{black}{\cite{stabilityai2024sd35}}:} 8B-parameter MMDiT with 38 layers and CLIP+T5 text encoders. FLOPs reflect quadratic attention scaling and joint text–image token processing. CFG is applied via 2× batch expansion. \textcolor{black}{\footnote{\url{https://huggingface.co/stabilityai/stable-diffusion-3.5-large}}}

\textbf{Flux.1 [dev] \textcolor{black}{\cite{blackforestlabs2024flux}}:} 12B-parameter hybrid MMDiT (19 layers + 38 single transformer) trained with rectified flow rather than standard diffusion. FLOP estimation adapts transformer attention for flow-matching dynamics. CFG incurs a 2× full model pass. \textcolor{black}{\footnote{\url{https://huggingface.co/black-forest-labs/FLUX.1-dev}}}

\textbf{Qwen-Image \textcolor{black}{\cite{wu2025qwenimagetechnicalreport}}:} 20B-parameter, 60-layer MMDiT with 16×16 visual patches and T5-based text conditioning. FLOP calculation follows the same MMDiT formulation used for Flux and SD3.5. CFG is implemented as a 2× forward pass. \textcolor{black}{\footnote{\url{https://huggingface.co/Qwen/Qwen-Image}}}

\textbf{Stable Diffusion 2 (SD2) \textcolor{black}{\cite{rombach2022high}}:} 865M-parameter U-Net baseline. FLOPs are decomposed across ResNet blocks, cross-attention layers, and skip connections. CFG is applied via batch-level duplication. \textcolor{black}{\footnote{\url{https://huggingface.co/stabilityai/stable-diffusion-2}}}

This selection spans 865M–20B parameters, convolutional vs. transformer architectures, diffusion vs. flow-matching objectives, and fixed- vs. variable-length text conditioning — enabling a robust evaluation of scaling law generalization across paradigms.

\subsection{Hardware Configuration}
Our experiments are conducted on three NVIDIA GPU architectures representing different performance and efficiency characteristics.

\textbf{NVIDIA A100:} A data center GPU with 80 GB of HBM2e memory, 6,912 CUDA cores, and 432 third-generation Tensor Cores. The A100 delivers state-of-the-art mixed-precision performance and serves as our primary experimental platform. Its peak theoretical throughput is 312 TFLOPS (bfloat16 / Tensor Core precision).

\textbf{NVIDIA RTX A4000:} A professional workstation GPU with $16$GB GDDR6 memory, $6144$ CUDA cores, and $192$ Tensor cores. This GPU represents mid-range deployment scenarios common in professional and edge applications. Peak theoretical performance: $19.2$ TFLOPS ($\text{fp32}$).

\textbf{NVIDIA RTX A6000 ADA:} A high-end workstation GPU with 48\,GB of GDDR6 ECC memory, 18{,}176 CUDA cores, and 568 Tensor Cores. This represents premium workstation deployment with substantial memory capacity and cutting-edge Ada Lovelace architecture. Peak theoretical performance: 91.1 TFLOPS (fp32).

Experiments are conducted using CUDA device isolation with individual GPU assignment per experiment to minimize resource contention. Energy measurements are performed during dedicated inference runs with consistent GPU synchronization to ensure power measurement accuracy.

\subsection{Hyperparameter Space}
Our experiments span a comprehensive hyperparameter space designed to capture the full range of practical deployment scenarios:

\begin{itemize}
    \item \textbf{Inference steps ($T$)}: $\{10, 20, 30, 40, 50\}$ - covering fast sampling to high-quality generation.
    \item \textbf{Image resolutions}: $\{256^2, 512^2, 768^2, 1024^2\}$ - from preview to high-resolution output. 
    \item \textbf{Precision}: $\{\text{float16}, \text{float32}\}$ - comparing memory and energy trade-offs.
    \item \textbf{Total queries ($N$)}: $\{25, 50, 100\}$ prompts - varying iterative inference throughput settings.
    \item \textbf{Classifier-free guidance}: $\{\text{enabled}, \text{disabled}\}$ - quality vs. efficiency trade-off.
\end{itemize}

This design yields 240 potential experimental configurations per model-GPU combination. However, practical limitations constrain our data collection: complete experimental sweeps are only achieved for A100 hardware, representing our most reliable platform. For other configurations (A4000/A6000 GPUs), we collect representative samples with varying levels of coverage across models. Additionally, Qwen's large memory requirements (\textcolor{black}{20B} parameters) prevent fp32 inference on our hardware, limiting this model to fp16 precision experiments. We prioritize data collection to ensure adequate coverage for cross-platform scaling validation while acknowledging these experimental constraints. 

\subsection{Dataset and Energy Units}
We utilize a random subset of the COCO 2017 dataset \cite{lin2015microsoftcococommonobjects} \textcolor{black}{, a widely adopted benchmark in text-to-image evaluation \cite{shih2023parallelsamplingdiffusionmodels, selvam2024selfrefiningdiffusionsamplersenabling}}, for our experiments
\footnote{\textcolor{black}{The prompts correspond to the first entries in \texttt{coco\_captions.txt} in our released codebase, curated from the open-source dataset available at \url{https://www.kaggle.com/datasets/awsaf49/coco-2017-dataset}.}}. \textcolor{black}{Since our energy measurements depend on architectural and inference parameters—such as resolution, step count, and precision—rather than dataset semantics, and since prompt embeddings are generally fixed-length, we do not expect substantial variation in energy consumption across datasets.} Energy consumption is recorded in kilowatt-hours (kWh) using our tracker. 
We model the natural logarithm of the total energy spent to generate $N$ prompts ($\log(E)$). We thus apply the exponential function on our assessment to obtain the estimated energy consumption ($E = \exp(\log(E))$ kWh). To convert to Joules, we apply the relation $E \text{ kWh} = E \times 3.6 \times 10^6 \text{ J}$.

\section{Results and Discussion}
\label{sec:results}
In this section, we evaluate the predictive power of our energy scaling laws across the experimental landscape. We begin by demonstrating high accuracy for individual models, followed by an assessment of robustness across heterogeneous GPU architectures and an exploration of cross-model generalization. For more details on FLOPs computations and energy contextualization, we refer the reader to Appendix~\ref{app:flops} and Appendix~\ref{app:appendix_energy}.

\subsection{Individual Model Energy Scaling}
\begin{figure}[t!]
    \centering
    \begin{subfigure}[b]{0.33\columnwidth}
        \centering
        \includegraphics[height=4.5cm]{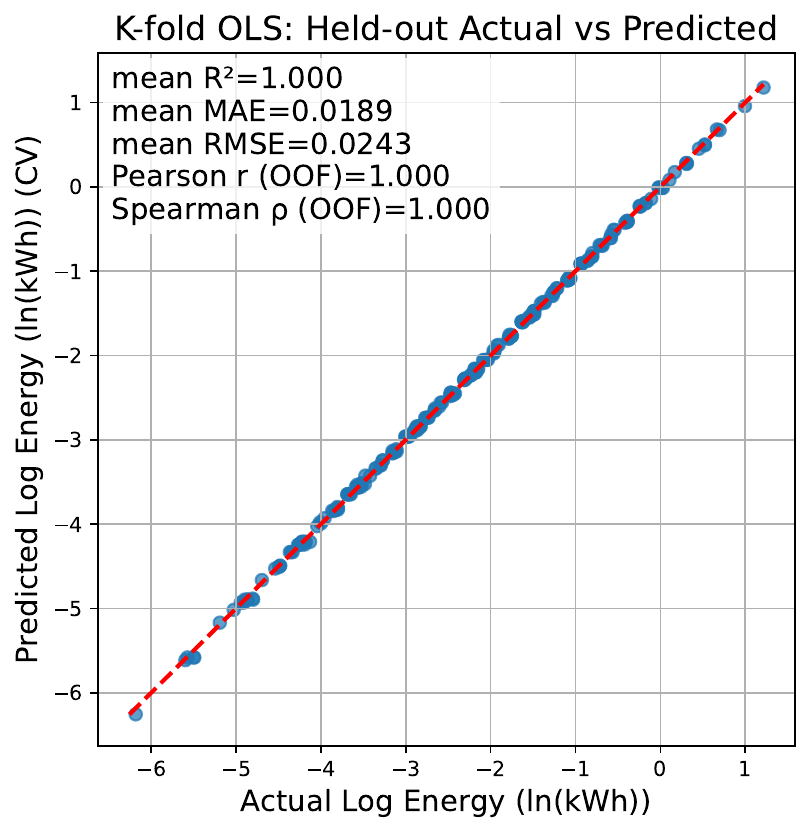}
        \caption{Flux}
        \label{fig:flux_actual_vs_predicted}
    \end{subfigure}
    \hfill
    \begin{subfigure}[b]{0.33\columnwidth}
        \centering
        \includegraphics[height=4.5cm]{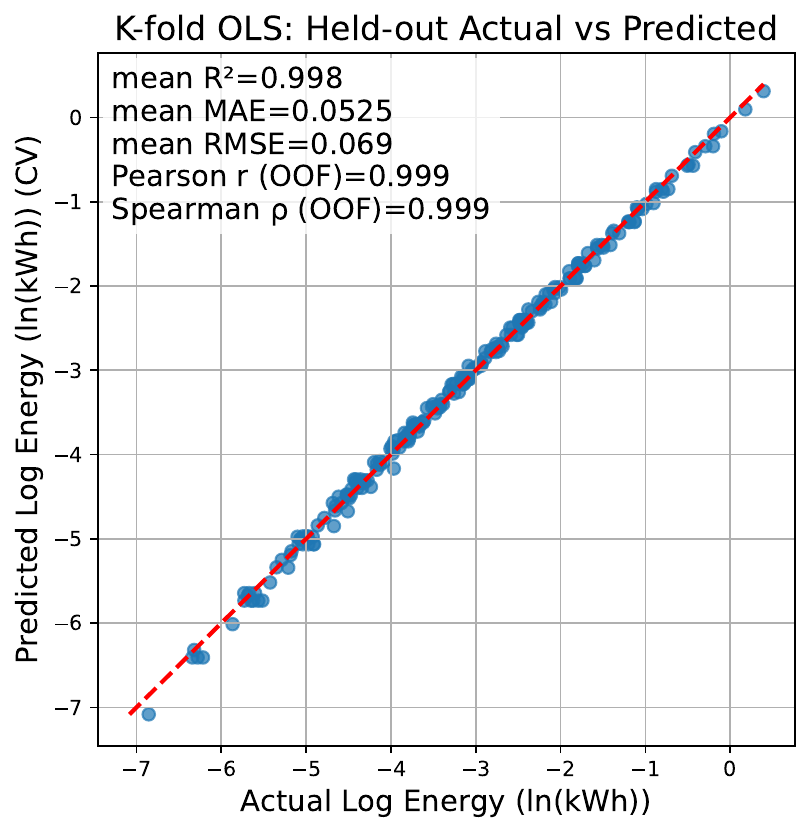}
        \caption{Stable Diffusion 3.5}
        \label{fig:sd35_actual_vs_predicted}
    \end{subfigure}
    \hfill
    \begin{subfigure}[b]{0.33\columnwidth}
        \centering
        \includegraphics[height=4.5cm]{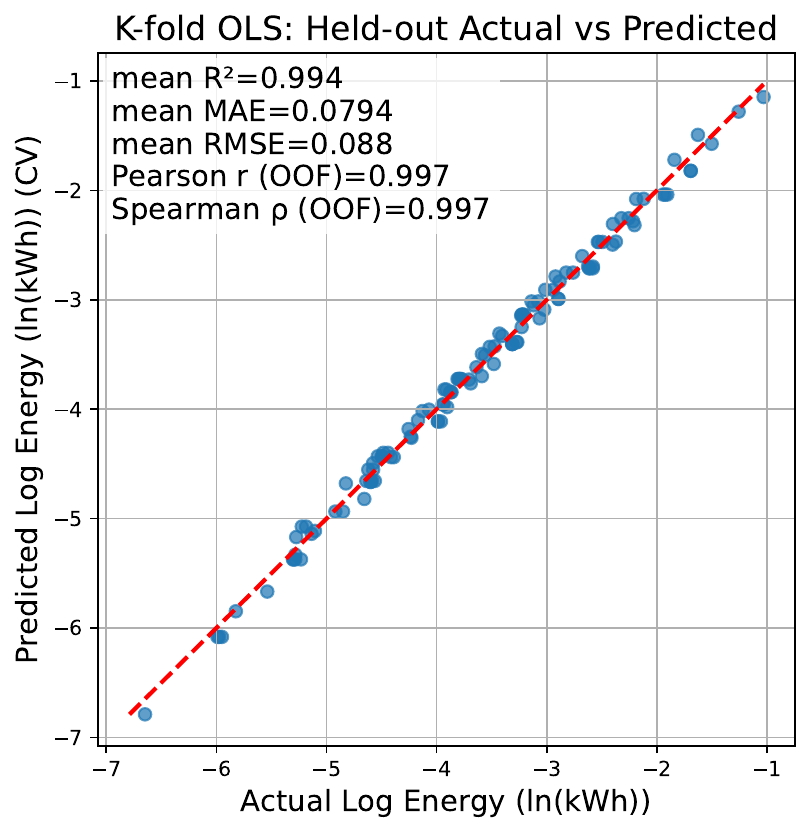}
        \caption{Qwen}
        \label{fig:qwen_actual_vs_predicted}
    \end{subfigure}
    \caption{Individual model energy scaling validation on NVIDIA A100 GPU. Diagnostic plots showing actual versus predicted energy consumption for (a) Flux, (b) Stable Diffusion 3.5, and (c) Qwen diffusion models.}
    \label{fig:model_specific_validation_plots}
\end{figure}

\textcolor{black}{Figure~\ref{fig:model_specific_validation_plots}} presents energy scaling validation for individual diffusion models on NVIDIA A100 hardware. All three models demonstrate strong adherence to our scaling law formulation with $R^2 > 0.92$, with learned parameters summarized in Table~\ref{tab:model_specific_scaling_parameters}.

\begin{table}[ht]
        \centering
        \adjustbox{height=1cm, center}{%
        \begin{tabular}{@{}lccc@{}}
            \toprule
            & \textbf{Flux} & \textbf{SD 3.5} & \textbf{Qwen} \\
            \midrule
            \textbf{$\log(A)$} & -20.61 & -19.95 & -18.64 \\
            \textbf{$\alpha$} & 0.989  & 0.969 & 0.992 \\
            \textbf{$\beta_{\text{dtype}}$} & 2.04 & 1.90 & 0 \\
            \textbf{$\beta_{\text{res}}$} & -0.043  & -0.054  & -0.306 \\
            \bottomrule
        \end{tabular}%
        }
        \vspace{0.3cm}
    \caption{The learned scaling parameters with exponents $\alpha$ approaching the theoretical compute-bound ideal of 1.0. All models exhibit near-linear FLOP-energy relationships, confirming the compute-bound nature of diffusion inference. Note: \textcolor{black}{$\beta_{\text{A4000}}$=$\beta_{\text{A6000}}$=0} since only A100 data is used; Qwen $\beta_{\text{dtype}}$=0 due to fp16-only training.}
    \label{tab:model_specific_scaling_parameters}
\end{table}

\textbf{Flux} achieves $R^2 = 1.0$ with scaling exponent $\alpha = 0.989$, very close to the theoretical compute-bound ideal of 1.0 \textcolor{black}{and confirming that diffusion inference is primarily compute-bound}. The precision coefficient $\beta_{\text{dtype}} = 2.04$ indicates approximately $\approx7.4 \times$ energy increase from FP16 to 32, possibly reflecting differences between NVIDIA’s optimized half-precision Tensor Cores and standard FP32 execution paths. We can also see that our energy range generally falls in $8.9\times10^3-9.8\times10^6J$.

\textbf{Stable Diffusion 3.5} shows similar performance with $R^2 = 0.998$ and $\alpha = 0.969$, demonstrating consistent scaling behavior despite different architectural details. \textcolor{black}{Moreover, this supports quadratic scaling of energy with spatial dimensions, as our FLOPs estimates following \cite{kaplan2020scaling} already incorporate a quadratic component in the number of tokens—consistent with the finding in \cite{delavande2025videokilledenergybudget}}. The predicted energy range across configurations spans $3.3\times10^3$–$9.8\times10^6$ J.

\textbf{Qwen} presents $R^2 = 0.994$ and $\alpha = 0.992$, however heavily using the $\beta_{\text{res}}$ bias, potentially reflecting memory bandwidth limitations in its 60-layer architecture. We can see that our energy range generally falls in $3.3\times10^3-1.32\times10^6J$, but only for float16 inferences. 

As a case study, Table~\ref{tab:energy_qwen} in the Appendix reports energy usage for Qwen across hyperparameter configurations on an A100 GPU in the 100-prompt setting. Consumption spans three orders of magnitude---from $1.83 \times 10^4$~J (0.051~Wh) per image under a minimal configuration (10 steps, $256^2$, fp16, no CFG) to $1.29 \times 10^6$~J (3.58~Wh) for high-quality generation (50 steps, $1024^2$, fp16, CFG). These values exceed typical large language model inference costs: a single diffusion image can consume up to 10$\times$ the energy of an average ChatGPT query (0.34~Wh)~\cite{samaltmanGentleSingularity} or median Gemini request (0.24~Wh)~\cite{elsworth2025measuring}.

Furthermore, the negative resolution-bias coefficient ($\beta_{\text{res}} = -0.306$ to $-0.043$) indicates that this bias diminishes as resolution increases. This suggests that GPU utilization becomes more efficient at larger tensor sizes—or, conversely, that fixed-overhead operations constitute a relatively larger share of the energy cost at lower resolutions.

\subsection{Cross-GPU Hardware Validation}

\begin{figure}[t!]
    \centering
    \begin{subfigure}[b]{0.45\columnwidth}
        \centering
        \includegraphics[height=5cm]{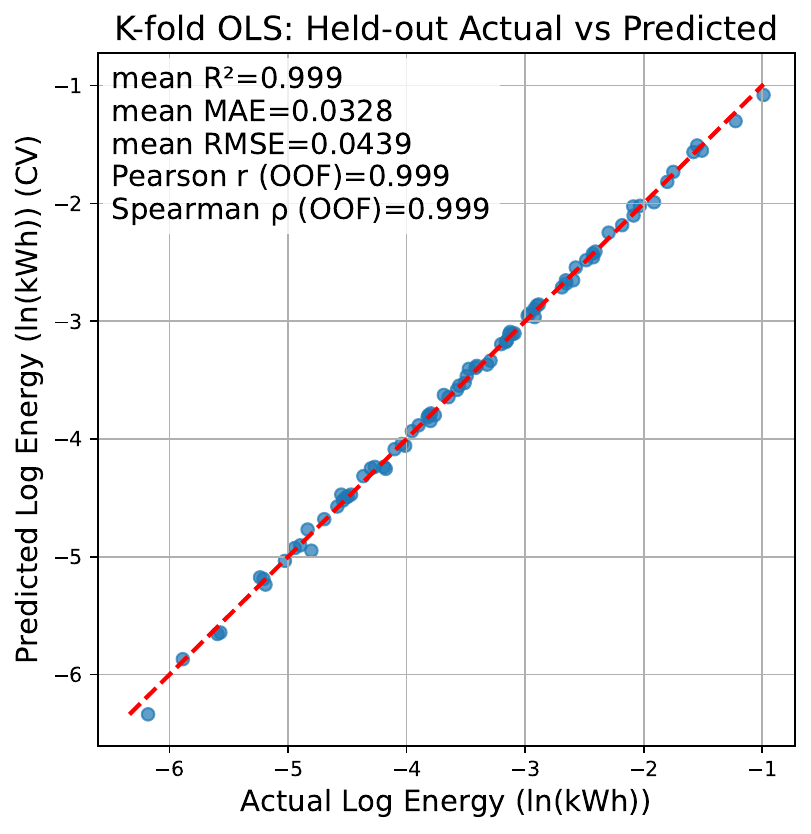}
        \Description{Actual vs predicted energy consumption for Flux model.}
        \caption{Flux}
        \label{fig:flux_actual_vs_predicted_cross}
    \end{subfigure}
    \hfill
    \begin{subfigure}[b]{0.45\columnwidth}
        \centering
        \includegraphics[height=5cm]{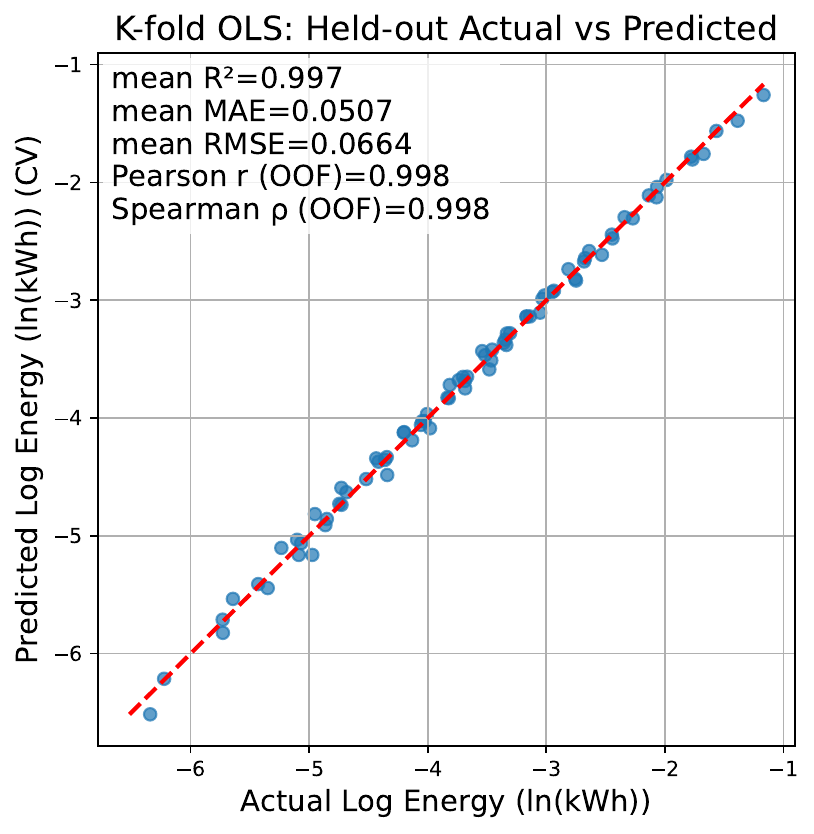}
        \caption{Stable Diffusion 3.5}
        \label{fig:sd35_actual_vs_predicted_cross}
    \end{subfigure}
    
    \caption{Diagnostic plots show actual versus predicted energy consumption for (a) Flux and (b) Stable Diffusion 3.5 models, using data from NVIDIA A100 and A6000 GPUs.}
    \label{fig:across_gpus_validation_plots}
\end{figure}

Plots in \textcolor{black}{Figure~\ref{fig:across_gpus_validation_plots}} demonstrates scaling law robustness across GPU architectures. The fundamental FLOPs scaling exponents remain stable ($\alpha = 0.997, 0.989$). GPU-specific coefficients in Table~\ref{tab:across_gpus_validation_parameters} capture the hardware differences: A6000 shows energy overhead ($\textcolor{black}{\beta_{\text{A6000}}} = 0.450, 0.308$) compared to A100 baseline. Qwen is excluded from cross-GPU validation due to memory constraints on the A6000 platform.

This hardware consistency validates our approach for deployment planning—organizations can predict energy consumption across different GPU platforms without extensive empirical testing for each configuration.

\begin{table}[ht]
    \centering
    \adjustbox{height=1.2cm, center}{%
    \begin{tabular}{@{}lccc@{}}
        \toprule
        & \textbf{Flux} & \textbf{SD 3.5} \\
        \midrule
        \textbf{$\log(A)$} & -20.85 & -20.44 \\
        \textbf{$\alpha$} & 0.997  & 0.989 \\
        \textcolor{black}{\textbf{$\beta_{\text{A6000}}$}} & 0.450 & 0.308 \\
        \textbf{$\beta_{\text{res}}$} & -0.027  & -0.037  \\
        \bottomrule
    \end{tabular}%
    }
    \vspace{0.3cm}
    \caption{The learned scaling parameters, highlighting stable exponents ($\alpha$) across GPUs and GPU-specific coefficients. Note: The Flux plot includes no-CFG, float16, and non-50-prompt runs, while the SD 3.5 plot includes only CFG, float16, and non-50-prompt runs.}
    \label{tab:across_gpus_validation_parameters}
\end{table}

\subsection{Cross-Model Generalization}

\begin{figure*}[ht]
    \centering
    
    \begin{subfigure}[b]{0.08\textwidth}
        \centering
        \rotatebox{90}{\textbf{Training}}
        \vspace{2cm}
    \end{subfigure}
    \begin{subfigure}[b]{0.29\textwidth}
        \centering
        \includegraphics[height=4cm]{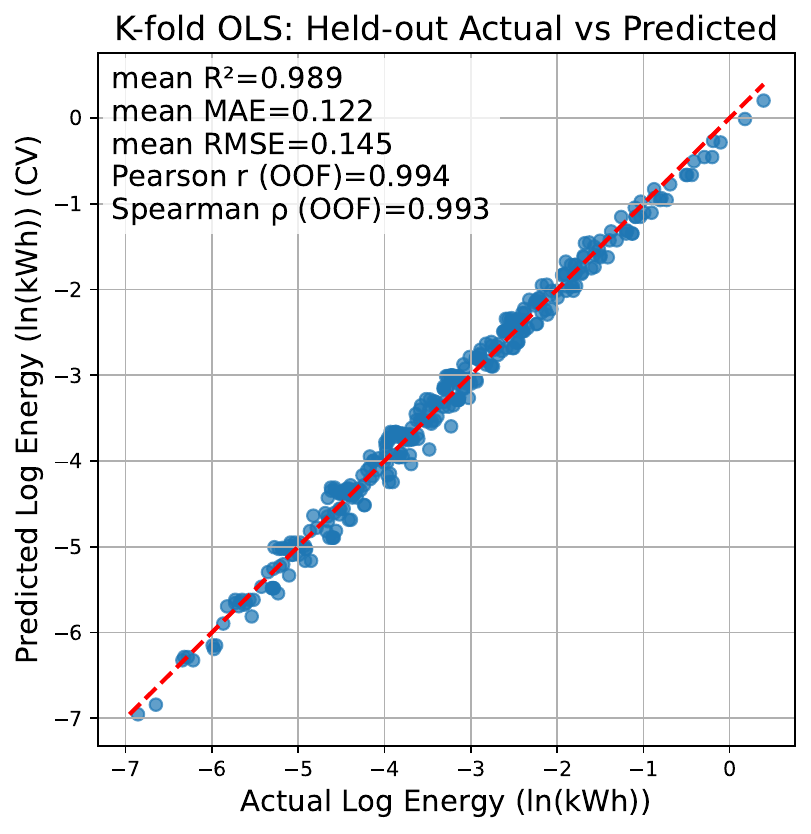}
        \caption{Qwen + Stable Diffusion 3.5}
        \label{fig:qwen_sd35_pair}
    \end{subfigure}
    \hfill
    \begin{subfigure}[b]{0.29\textwidth}
        \centering
        \includegraphics[height=4cm]{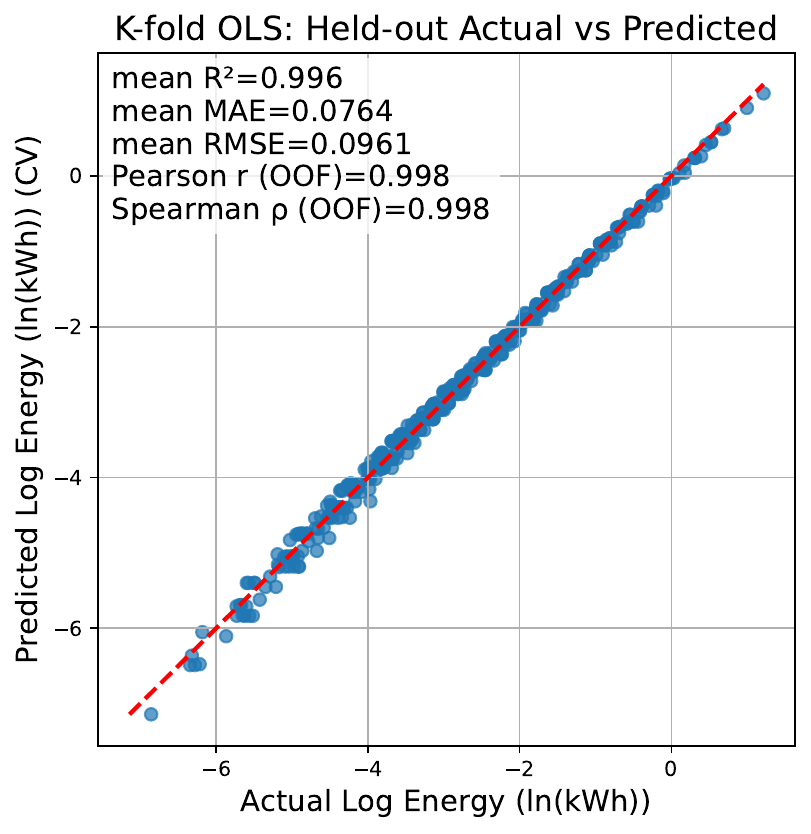}
        \caption{Flux + Stable Diffusion 3.5}
        \label{fig:flux_sd35_pair}
    \end{subfigure}
    \hfill
    \begin{subfigure}[b]{0.29\textwidth}
        \centering
        \includegraphics[height=4cm]{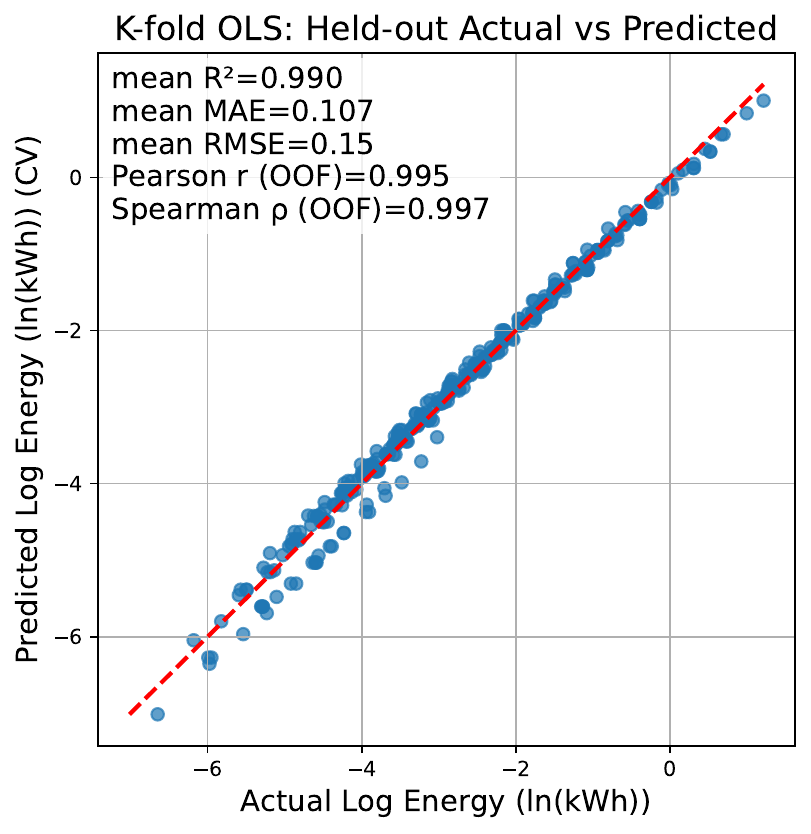}
        \caption{Flux + Qwen}
        \label{fig:flux_qwen_pair}
    \end{subfigure}
    
    \vspace{0.3cm}
    
    \begin{subfigure}[b]{0.08\textwidth}
        \centering
        \rotatebox{90}{\textbf{Testing}}
        \vspace{2cm}
    \end{subfigure}
    \begin{subfigure}[b]{0.29\textwidth}
        \centering
        \includegraphics[height=4cm]{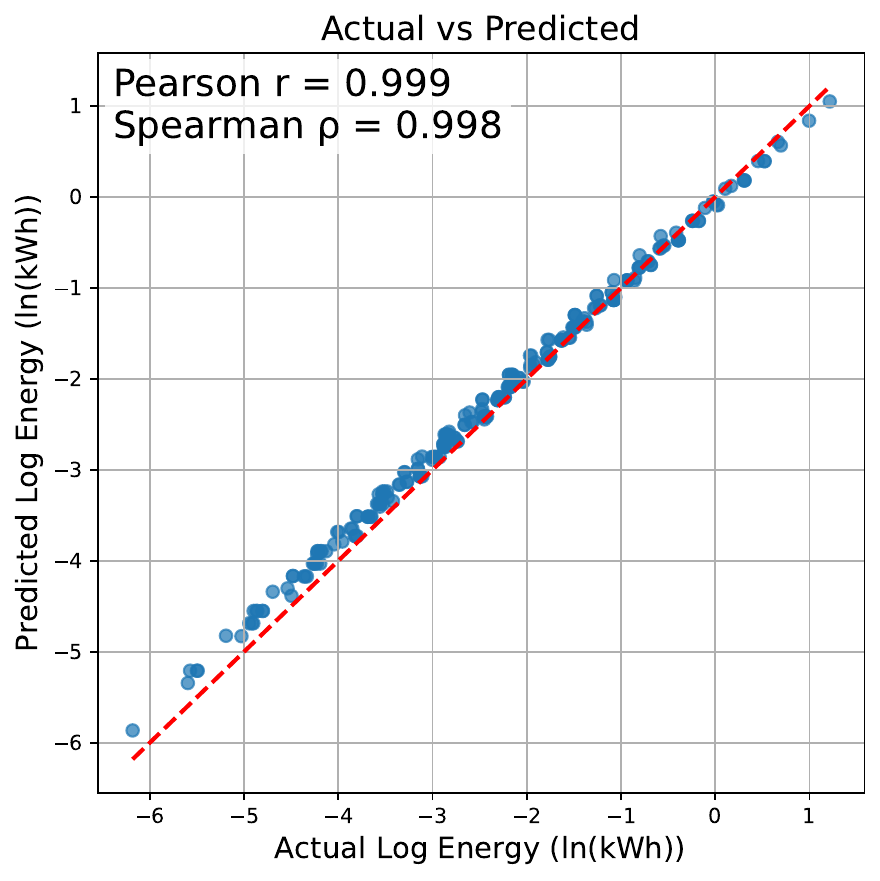}
        \caption*{Flux}
        \label{fig:flux_test}
    \end{subfigure}
    \hfill
    \begin{subfigure}[b]{0.29\textwidth}
        \centering
        \includegraphics[height=4cm]{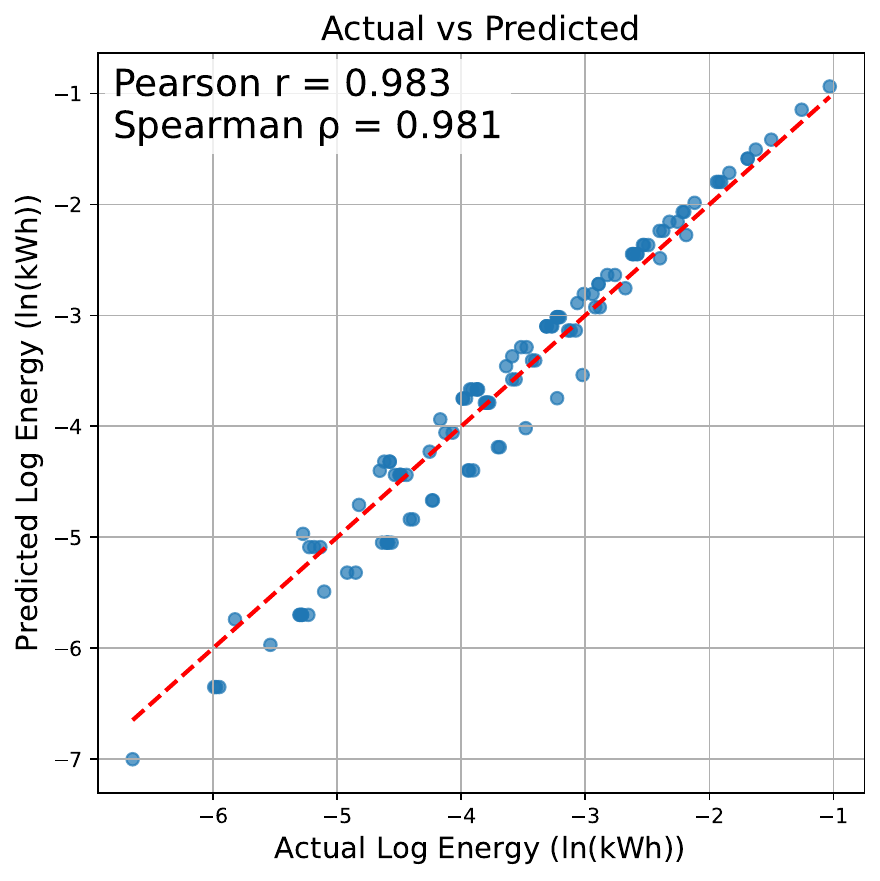}
        \caption*{Qwen}
        \label{fig:qwen_test}
    \end{subfigure}
    \hfill
    \begin{subfigure}[b]{0.29\textwidth}
        \centering
        \includegraphics[height=4cm]{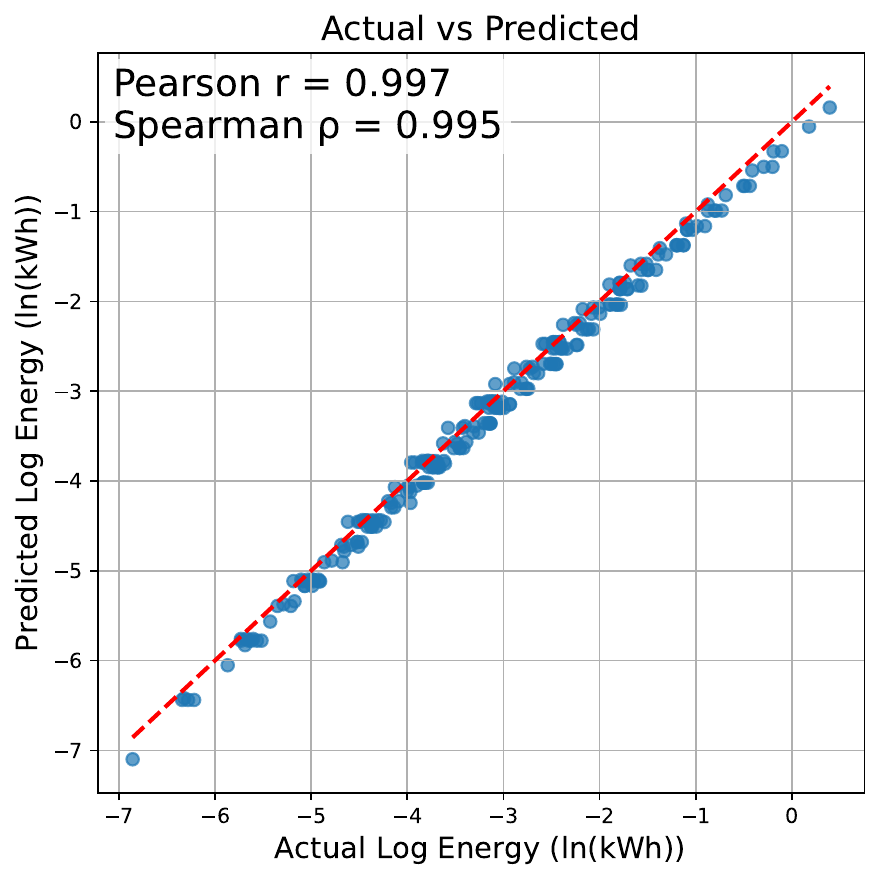}
        \caption*{Stable Diffusion 3.5}
        \label{fig:sd35_test}
    \end{subfigure}
    
    \caption{The top row shows training on model pairs: (a) Qwen + SD 3.5, (b) Flux + SD 3.5, and (c) Flux + Qwen. The bottom row presents the corresponding tests on the held-out models: Flux, Qwen, and SD 3.5, respectively. Consistent diagnostic patterns across all training–test pairs demonstrate the robustness of our FLOP-based scaling methodology for cross-model energy prediction. All plots here use results from the NVIDIA A100, which yielded the most comprehensive hyperparameter search.}
    \label{fig:cross_model_validation}
\end{figure*}

The most significant finding is demonstrated in \textcolor{black}{Figure~\ref{fig:cross_model_validation}}: scaling laws learned from certain diffusion models successfully predict energy consumption for entirely different diffusion models. This cross-model universality suggests that energy scaling laws capture fundamental computational principles rather than model-specific optimizations, enabling transferability from open-source models to proprietary closed systems, with profound implications for sustainable AI development.

Three cross-validation scenarios were used to assess generalization: (a) Qwen + SD 3.5 → Flux, (b) Flux + SD 3.5 → Qwen, and (c) Flux + Qwen → SD 3.5. Strong agreement between training and testing performance indicates that energy efficiency is governed primarily by computational complexity rather than architectural hyperparameters. The slight deviation observed in the Qwen cases stems from not fitting the $\beta_{\text{res}}$ parameter. We find that the off-diagonal points correspond to underestimated FLOPs for $256\times256$ images. At this smaller image resolution, we hypothesize hat FLOPs alone no longer significantly dominate the full energy costs, relative to memory operations and bandwidth constraints. Consequently, the bounding characteristics of FLOPs become insufficient to explain the observed energy profile in large models. The  $\beta_{\text{res}}$ parameter likely adjusts for this when fitted on the Qwen data.

\subsection{Cross-Architecture (\& GPU) Validation}
We further validate the transferability of our scaling laws across fundamentally different architectural paradigms, including convolutional U-Nets (Stable Diffusion 2) and transformer-based MMDiT models. Evaluations conducted on A100 and A6000 show that training on MMDiT models (Flux, SD 3.5, Qwen) and testing on the U-Net architecture achieves robust relative ranking performance ($R, R_s > 0.9$), although decrease in precision. 

This cross-architecture generalization demonstrates that FLOP-based scaling laws capture computational universals rather than architecture-specific optimizations, enabling unified energy prediction across diverse model families from convolutional to transformer designs (detailed results in Appendix~\ref{sec:appendix}).

\begin{figure}[ht]
    \centering
    
    \begin{subfigure}[b]{0.45\columnwidth}
        \centering
        \includegraphics[height=5cm]{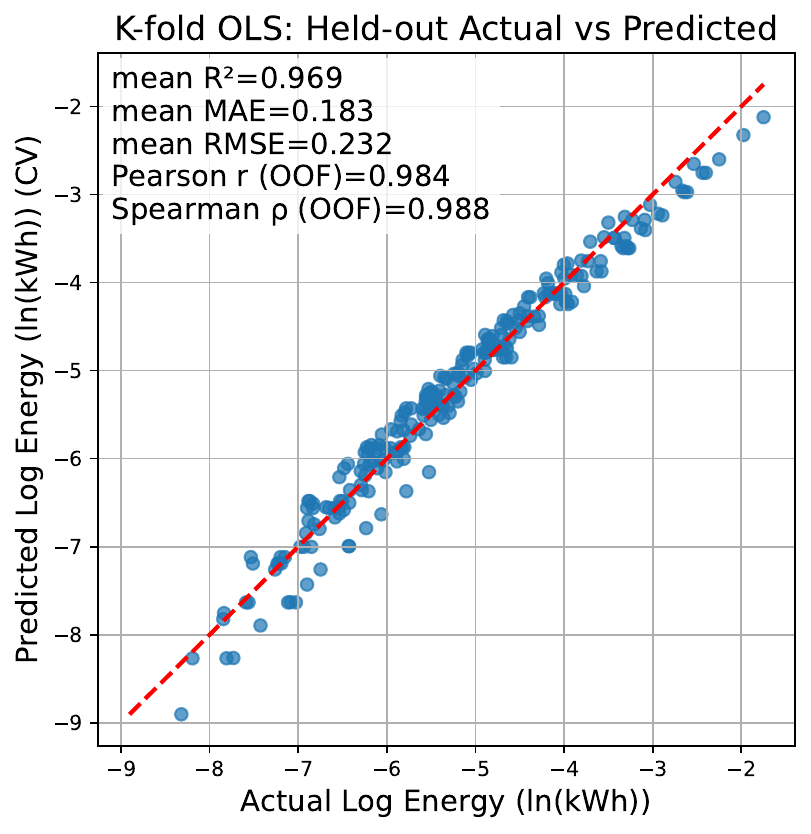}
        \caption{Stable Diffusion 2 (NVIDIA A100)}
        \label{fig:sd2_a100}
    \end{subfigure}
    \hfill
    \begin{subfigure}[b]{0.45\columnwidth}
        \centering
        \includegraphics[height=5cm]{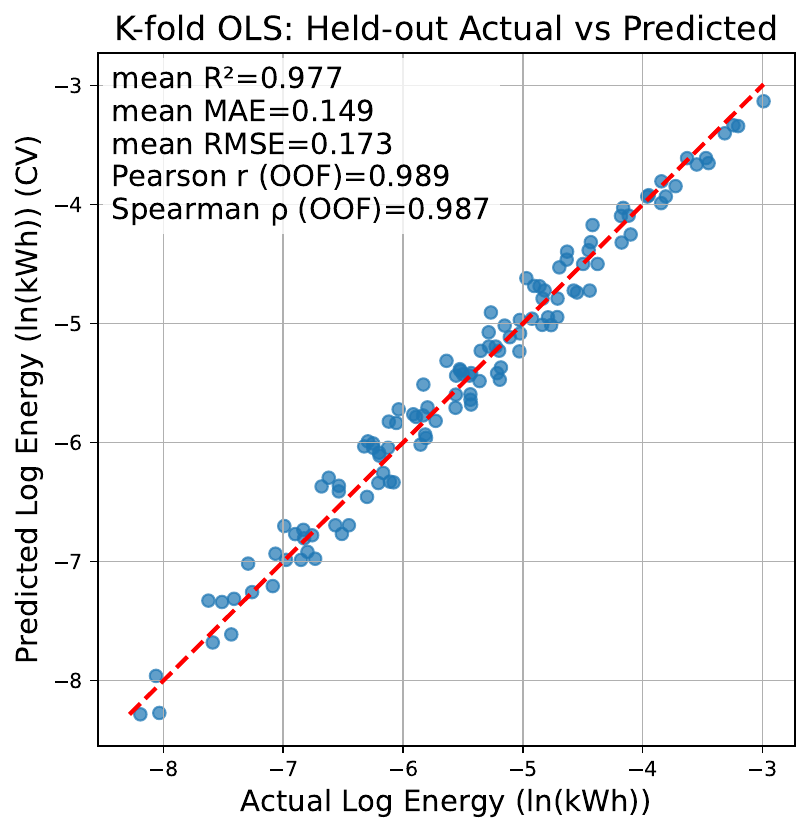}
        \caption{Stable Diffusion 2 (cross-GPU)}
        \label{fig:sd2_cross_gpu}
    \end{subfigure}
     
    \caption{Diagnostic plots show (a) Stable Diffusion 2 on NVIDIA A100, illustrating U-Net scaling behavior, and (b) cross-GPU validation across A100, A4000, and A6000 platforms. Consistent scaling patterns confirm that our FLOP-based energy prediction generalizes beyond transformer-based models to convolutional architectures. Note: Cross-GPU results include CFG, float16, and non-50-prompt runs.}
    \label{fig:individual_non_mmdit_architecture_validation}
\end{figure}

\subsection{Individual U-Net Architecture Validation}

\textcolor{black}{Figure~\ref{fig:individual_non_mmdit_architecture_validation}} provides detailed individual model validation for Stable Diffusion 2's U-Net architecture across different GPU platforms. Despite fundamental architectural differences from transformer-based models, SD2 demonstrates robust scaling behavior consistent with MMDiT architectures, with $R^2 > 0.9$ for both single-GPU and cross-GPU validation scenarios. See Table \ref{tab:sd2_details} for the SD2 FLOP details.

\subsection{U-Net to Transformer Generalization}
Figure~\ref{fig:cross_non_mmdit_architecture_validation} in the Appendix shows that cross-architecture experiments accurately predict U-Net energy consumption using scaling laws learned from transformer-based MMDiT models, and vice versa. Figure~\ref{fig:cross_non_mmdit_architecture_gpu_validation} (also in the Appendix) further demonstrates this generalization across A100 and A6000 GPUs.

Despite the architectural gulf between U-Net and transformer designs, our scaling laws capture fundamental energy-complexity relationships independent of specific architectural paradigms, successfully bridging traditional convolutional U-Net designs and modern transformer-based MMDiT approaches. This indicates broad applicability across diffusion model paradigms and suggests our methodology could extend to other generative model families.

\begin{figure}[ht]
    \centering
    \includegraphics[width=0.98\linewidth]{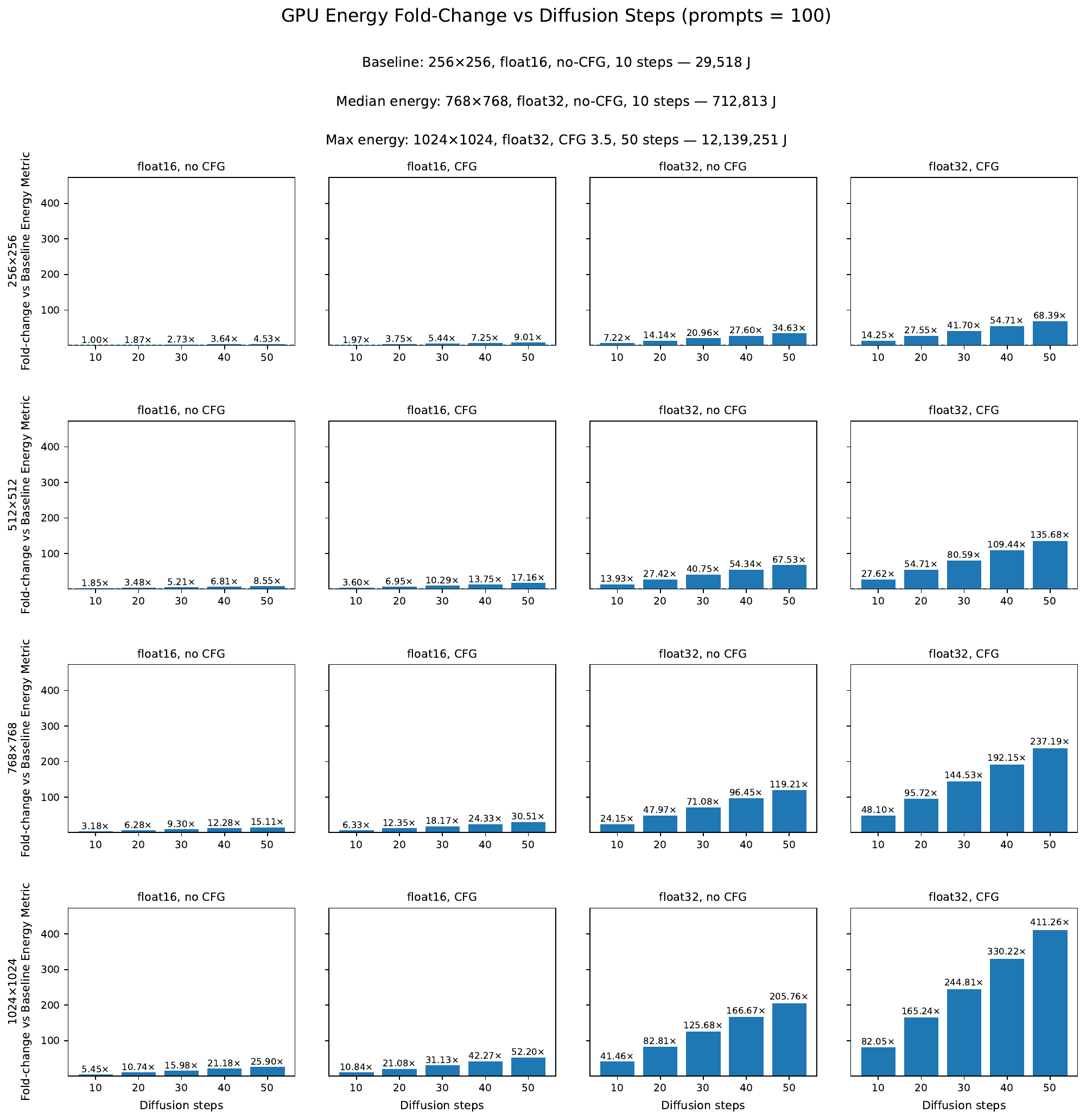}
    \caption{
        Energy scaling of Flux, relative to smallest energy setting (baseline). Flux has 38 layers in its MMDiT and 17B parameters total (12B MMDit, 5B text encoder, 80M VAE parameters).
    }
    \label{fig:flux_energy_scale}
\end{figure}

\subsection{Hyperparameter Energy Contributions}
\label{sec:energy_context}

In Figure \ref{fig:flux_energy_scale}, we compare how energy consumption scales of Flux as a function of key hyperparameters, normalized to the lowest-energy configuration in the 100-prompt A100 setting: $256 \times 256$ resolution, float16 precision, no CFG, and 10 diffusion steps. The highest-energy setting corresponds to $1024 \times 1024$ resolution, float32 precision, CFG enabled, and 50 steps. Across all models, float32 incurs significantly higher energy cost than float16, and energy increases sharply with image resolution — especially when paired with more denoising steps. These observations mirror recent efforts toward step-efficient and distillation-based generation strategies to reduce inference cost at scale. Exact A100 energy values for these configurations are reported in the Appendix (Tables~\ref{tab:energy_flux}--\ref{tab:energy_qwen}). 

\subsection{Practical Implications and Applications}

Our findings support several immediate applications for sustainable AI deployment:

\textbf{Deployment and Hardware Planning.} Near-linear FLOPs scaling ($\alpha \approx 1$) enables accurate pre-deployment energy estimates, allowing practitioners to compare GPU configurations, balance cost–efficiency trade-offs, and select deployment sites to minimize environmental impact.

\textbf{Carbon-Aware and Algorithmic Optimization.} Consistent hyperparameter effects allow systems to dynamically adjust precision, resolution, or step count based on grid carbon intensity, while our framework also quantifies the sustainability gains from algorithmic improvements such as reduced inference steps or more efficient sampling.

\textbf{Standardization and Reporting.} The proposed methodology offers a standardized approach for estimating and reporting energy consumption, supporting regulatory compliance and reproducible energy accounting across models and hardware platforms.

\subsection{Limitations and Future Directions}

While our results show strong predictive performance, several limitations remain:

\textbf{Model and Hardware Scope.} We evaluate four models and three NVIDIA GPU types. New diffusion architectures and alternative hardware (e.g., AMD, specialized accelerators) may exhibit distinct scaling, though cross-architecture validation indicates reasonable robustness. \textcolor{black}{Moreover, while using CodeCarbon for estimating energy consumption is a straightforward approach, this however makes pragmatic assumptions that should be recognized.}

\textbf{System-Wide Scope.} Our analysis focuses exclusively on dynamic GPU energy consumption, as it represents the dominant variable cost for diffusion-based inference. However, a full system-wide assessment would also need to account for the energy usage of the CPU, RAM, and storage, as well as the overhead of data center cooling systems and networking. \textcolor{black}{Prior work has also shown that that commonly used compute proxies such as parameter count or FLOPs may not correlate with actual runtime or efficiency due to factors such as parallelism, memory access patterns, and hardware-specific optimizations \cite{dehghani2022efficiencymisnomer}. In addition, while experiments are designed to evaluate inference in interactive deployment scenarios, batch sizes can influence GPU utilization and alter the system-level energy profiles. Future work could extend this analysis to examine how batching affects the scaling relationships observed in this study.}

\textbf{Dynamic Effects.} GPU frequency scaling and thermal throttling may affect real-world energy use beyond what static FLOP-based models capture, despite our controlled experimental setup.

\textbf{Pipeline Coverage.} Our analysis isolates the core model inference cost and does not account for auxiliary pre/post-processing or multi-model orchestration, which are typically minor contributors to overall energy. In real-world deployment, inference is also often served as part of a pipeline of user queries, which we approximate via iterative prompt evaluation, though this is not executed within a fully productionized serving stack.

\textcolor{black}{\textbf{Quality--Energy Trade-offs.} Deployment decisions often require balancing output quality against computational cost, as fidelity typically increases with the number of denoising steps or guidance strength—both of which raise energy consumption. This represents a Pareto optimization problem requiring nuanced approaches for assessing, comparing, and reporting such trade-offs \cite{10.1145/3715275.3732006, fischer_dami}. Our energy estimates can be combined with task-specific quality metrics to construct quality--energy Pareto frontiers for particular models and applications—for instance, by characterizing how denoising step count jointly affects output fidelity and energy demand—and techniques such as diffusion distillation \cite{salimans2022progressive, yin2024onestepdiffusiondistributionmatching} further illustrate how model design can shift these trade-offs by improving output quality at a fixed inference budget.}

Future work will extend our analysis to video diffusion models and those with larger scale, real-time carbon-aware optimization, \textcolor{black}{calibration of energy-quality Pareto frontiers}, and validation on broader hardware including emerging AI accelerators.

\section{Conclusion}
\label{sec:conclusion}
This work presents a principled framework for predicting diffusion model energy consumption from computational complexity. By adapting Kaplan-style scaling laws to energy prediction, we establish a theoretical foundation and a predictive methodology for sustainable AI deployment.

Our main contributions are: (1) a theoretical formulation of energy scaling laws incorporating computational, hardware, and resolution effects; (2) comprehensive validation across four model architectures and three GPU platforms; (3) demonstration of cross-architecture generalization enabling prediction for unseen model–hardware pairs; and (4) practical applications for deployment planning and carbon accounting. We find that diffusion model energy consumption can span three orders of magnitude (e.g., for Qwen: 0.051–3.58 Wh per image), with a single high-quality image consuming up to 10× the energy of a typical large language model query—underscoring the need for energy-aware deployment strategies.

Observed near-linear scaling confirms that diffusion inference is largely compute-bound, while robust cross-architecture transfer shows that these laws capture fundamental, architecture-agnostic principles. The framework achieves strong predictive accuracy ($R^2 > 0.9$ within-architecture, ($R, R_s > 0.9$) cross-architecture) and yields interpretable insights into hardware efficiency.

Beyond research, our approach supports sustainable AI practices through deployment planning, carbon-aware optimization, and standardized energy reporting. More broadly, it illustrates how scaling-law methodologies can generalize across models and hardware to guide systematic environmental optimization in machine learning.

\section*{Generative AI Usage Statement}
The authors used generative AI tools to assist with the formatting and to improve the grammar and fluency of the manuscript.

\clearpage 

\bibliographystyle{ACM-Reference-Format}
\bibliography{references}

\appendix

\section{Appendix}
\label{sec:appendix}
\subsection{FLOPs Computations}
\label{app:flops}
Tables \ref{tab:sd2_details}, \ref{tab:flop_formulas} and \ref{tab:model_flops_breakdown} detail how the FLOPs are computed per model class. While we detail the computation of the encoder and decoder flops, this is largely not used in our analysis due to the bounding behavior of the denoising computation. 

\begin{table*}[ht!]
\centering
\begin{minipage}{\linewidth}
\small
\centering \textbf{Stable Diffusion 2 (SD2) Inference Complexity Breakdown} \vspace{0.5em}

\noindent \textit{Conventions}: $h_0 = H/8$, $w_0 = W/8$. 2 FLOPs per mult-add. Totals in \textbf{GFLOPs} ($10^9$). Bespoke biases omitted for brevity.

\vspace{0.4em}
\begin{adjustbox}{max width=\linewidth}
\renewcommand{\arraystretch}{1.1}
\setlength{\tabcolsep}{6pt}
\begin{tabular}{lll}
\toprule
\textbf{Stage} & \textbf{Resolution} & \textbf{FLOP Components} \\
\midrule
ConvIn & $h_0 \times w_0$ & 
$\text{FLOPs}_{\text{conv}}(3, 4, 320, h_0, w_0)$ \\
\hline
\\[-0.6em]
Down 1 & $h_0{\times}w_0 \to \frac{h_0}{2}{\times}\frac{w_0}{2}$ & 
$2 \text{Res}(3, 320, 320, h_0, w_0) + 2 \text{Tr}(320, 0, 320, 1, h_0 w_0) + 2 \text{CrossTr}(320, 1024, 1920, 320, 1, h_0 w_0, 77) + \text{conv}(3, 320, 320, \frac{h_0}{2}, \frac{w_0}{2}) + \dots$ \\
\hline
\\[-0.6em]
Down 2 & $\frac{h_0}{2}{\times}\frac{w_0}{2} \to \frac{h_0}{4}{\times}\frac{w_0}{4}$ & 
$\text{Res}(3, 320, 640) + \text{Res}(3, 640, 640) + 2 \text{Tr}(640, \dots) + 2 \text{CrossTr}(640, \dots) + \text{conv}(3, 640, 640, \frac{h_0}{4}, \frac{w_0}{4})$ \\
\hline
\\[-0.6em]
Down 3 & $\frac{h_0}{4}{\times}\frac{w_0}{4} \to \frac{h_0}{8}{\times}\frac{w_0}{8}$ & 
$\text{Res}(3, 640, 1280) + \text{Res}(3, 1280, 1280) + 2 \text{Tr}(1280, \dots) + 2 \text{CrossTr}(1280, \dots) + \text{conv}(3, 1280, 1280, \frac{h_0}{8}, \frac{w_0}{8})$ \\
\hline
\\[-0.6em]
Down 4 & $h_0/8 \times w_0/8$ & 
$2 \cdot \text{FLOPs}_{\text{Res}}(3, 1280, 1280, h_0/8, w_0/8)$ \\
\hline
\hline
\\[-0.6em]
Mid & $h_0/8 \times w_0/8$ & 
$2 \text{Res}(3, 1280, 1280) + \text{Tr}(1280, 0, 1280, 1, \frac{h_0 w_0}{64}) + \text{CrossTr}(1280, 1024, 7680, 1280, 1, \frac{h_0 w_0}{64}, 77) + \dots$ \\
\hline
\hline
\\[-0.6em]
Up 1 & $h_0/8{\times}w_0/8 \to h_0/4{\times}w_0/4$ & 
$3 \text{Res}(3, 2560, 1280, h_0/8, w_0/8) + \text{conv}(3, 1280, 1280, h_0/4, w_0/4)$ \\
\hline
\\[-0.6em]
Up 2 & $h_0/4{\times}w_0/4 \to h_0/2{\times}w_0/2$ & 
$2 \text{Res}(3, 2560, 1280) + \text{Res}(3, 1920, 1280) + 3 \text{Tr}(1280, \dots) + 3 \text{CrossTr}(1280, \dots) + \text{conv}(3, 1280, 1280, \frac{h_0}{2}, \frac{w_0}{2})$ \\
\hline
\\[-0.6em]
Up 3 & $h_0/2{\times}w_0/2 \to h_0{\times}w_0$ & 
$\text{Res}(3, 1920, 640) + \text{Res}(3, 1280, 640) + \text{Res}(3, 960, 640) + 3 \text{Tr}(640, \dots, \frac{h_0 w_0}{4}) + 3 \text{CrossTr}(640, \dots) + \text{conv}(3, 640, 640, h_0, w_0)$ \\
\hline
\\[-0.6em]
Up 4 & $h_0 \times w_0$ & 
$2 \text{Res}(3, 640, 320) + \text{Res}(3, 960, 320) + 3 \text{Tr}(320, \dots) + 3 \text{CrossTr}(320, \dots) + \text{proj\_in/out}$ \\
\hline
\\[-0.6em]
ConvOut & $h_0 \times w_0$ & 
$\text{FLOPs}_{\text{conv}}(3, 320, 4, h_0, w_0)$ \\
\bottomrule
\end{tabular}
\end{adjustbox}
\end{minipage}
\caption{Architectural FLOP decomposition for Stable Diffusion 2 (SD2) on a per-step basis. Components are simplified for compactness (derived from Table~\ref{tab:model_flops_breakdown}).}
\label{tab:sd2_details}
\end{table*}

\begin{table*}[ht]
\centering
\caption{Summary of FLOP formulas used for different model components. Multiply–add pairs count as 2 FLOPs. Totals are for a single forward pass.}
\label{tab:flop_formulas}
\small
\resizebox{\textwidth}{!}{%
\begin{tabular}{ll}
\toprule
\textbf{Function} & \textbf{Formula} \\
\midrule
\textbf{Convolution} &
$\text{FLOPs}_{\text{conv}}(\text{kernel}=k, C_{\text{in}}, C_{\text{out}}, H, W) = 2 H W k^2 C_{\text{in}} C_{\text{out}}$
\\[0.5em]
\hline
\\[-0.5em]
\textbf{Transformer} &
$\begin{aligned}
&\text{FLOPs}_{\text{Tr}}(d_{\text{model}}, d_{\text{ff}}, d_{\text{attn}}, n_{\text{layer}}, n_{\text{ctx}}) \\
P &= 2 d_{\text{model}} n_{\text{layer}} (2 d_{\text{attn}} + d_{\text{ff}}),\quad
c_{\text{fwd}} = 2P + 2 n_{\text{layer}} n_{\text{ctx}} d_{\text{attn}} \\
\text{FLOPs}_{\text{Tr}} &= n_{\text{ctx}} \cdot c_{\text{fwd}}
\end{aligned}$
\\[0.5em]
\hline
\\[-0.5em]
\textbf{MMDiT} &
$\begin{aligned}
&\text{FLOPs}_{\text{MMDiT}}(d_{\text{model}}, d_{\text{ff}}, d_{\text{attn}}, n_{\text{layer}}, n_{\text{ctx}}) \\
P &= 4 d_{\text{model}} n_{\text{layer}} (2 d_{\text{attn}} + d_{\text{ff}}),\quad
c_{\text{fwd}} = P + 2 n_{\text{layer}} n_{\text{ctx}} d_{\text{attn}} \\
\text{FLOPs}_{\text{MMDiT}} &= n_{\text{ctx}} \cdot c_{\text{fwd}}
\end{aligned}$
\\[0.5em]
\hline
\\[-0.5em]
\textbf{Cross-Attn Tr.} &
$\begin{aligned}
&\text{FLOPs}_{\text{CrossTr}}(d_q, d_k, d_{\text{ff}}, d_{\text{attn}}, n_{\text{layer}}, n_q, n_k) \\
P &= 2 n_{\text{layer}} [d_q (d_{\text{attn}} + d_{\text{ff}}) + d_k d_{\text{attn}}],\quad
c_{\text{fwd}} = 2P + 2 n_{\text{layer}} n_k d_{\text{attn}} \\
\text{FLOPs}_{\text{CrossTr}} &= n_q \cdot c_{\text{fwd}}
\end{aligned}$
\\[0.5em]
\hline
\\[-0.5em]
\textbf{ResNetBlock} &
$\begin{aligned}
&\text{FLOPs}_{\text{Res}}(\text{kernel}=k, C_{\text{in}}, C_{\text{out}}, H, W) \\
&= \text{FLOPs}_{\text{conv}}(k, C_{\text{in}}, C_{\text{out}}, H, W) + \text{FLOPs}_{\text{conv}}(k, C_{\text{out}}, C_{\text{out}}, H, W)
\end{aligned}$
\\[0.5em]
\hline
\\[-0.5em]
\textbf{Decoder} &
$\begin{aligned}
&\text{FLOPs}_{\text{Dec}}(H, W) \text{ where } H_0 = H/8,\, W_0 = W/8 \\
&= \text{FLOPs}_{\text{conv}}(k{=}3, C_{\text{in}}{=}16, C_{\text{out}}{=}512, H_0, W_0) 
+ 2\text{FLOPs}_{\text{Res}}(k{=}3, 512, 512, H_0, W_0)
+ \text{FLOPs}_{\text{Tr}}(d_{\text{model}}{=}512, d_{\text{ff}}{=}256, d_{\text{attn}}{=}512, n_{\text{layer}}{=}1, n_{\text{ctx}}{=}H_0 W_0) \\
&\quad+ 3\text{FLOPs}_{\text{Res}}(3, 512, 512, H_0, W_0) 
+ \text{FLOPs}_{\text{conv}}(3, 512, 512, 2H_0, 2W_0) 
+ 3\text{FLOPs}_{\text{Res}}(3, 512, 512, 2H_0, 2W_0) 
+ \text{FLOPs}_{\text{conv}}(3, 512, 512, 4H_0, 4W_0) \\
&\quad+ \text{FLOPs}_{\text{Res}}(3, 512, 256, 4H_0, 4W_0) 
+ 2\text{FLOPs}_{\text{Res}}(3, 256, 256, 4H_0, 4W_0) 
+ \text{FLOPs}_{\text{conv}}(3, 256, 256, 8H_0, 8W_0) 
+ \text{FLOPs}_{\text{Res}}(3, 256, 128, 8H_0, 8W_0) \\
&\quad+ 2\text{FLOPs}_{\text{Res}}(3, 128, 128, 8H_0, 8W_0) 
+ \text{FLOPs}_{\text{conv}}(3, 128, 3, 8H_0, 8W_0)
\end{aligned}$
\\
\bottomrule
\end{tabular}%
}
\end{table*}

\begin{table*}[ht]
\centering
\caption{
Per-model FLOP composition (single forward pass). 
Multiply–add pairs count as 2 FLOPs. 
Totals are expressed in GFLOPs as defined in Table~\ref{tab:flop_formulas}.
Model-specific latent and embedding dimensions are substituted directly into
the general formulas.
}
\label{tab:model_flops_breakdown}
\setlength{\tabcolsep}{6pt}
\renewcommand{\arraystretch}{1.2}
\resizebox{\textwidth}{!}{%
\begin{tabular}{lll}
\toprule
\textbf{Model} & \textbf{Context / Geometry} & \textbf{FLOP formula (GFLOPs)} \\
\midrule
\textbf{FLUX} &
$\displaystyle n_{\text{ctx}}=\frac{HW}{16^2}+512$ &
\begin{minipage}[t]{0.64\linewidth}\small
Latent patch tokens ($H/16\times W/16$) plus 512 fixed text tokens.  
Hidden width $d_{\text{model}}{=}3072$, $d_{\text{ff}}{=}12288$, $d_{\text{attn}}{=}3072$. 
Uses 19 MMDiT layers and 38 transformer layers. Text encoders: OpenAI CLIP and T5. 
The T5 bias term accounts for GLU activation blocks.
\\[4pt]
\textit{Core (diffusion):}\\
$\displaystyle 
\text{GFLOPs}_{\text{MMDiT}}(3072,12288,3072,19,n_{\text{ctx}})
+
\text{GFLOPs}_{\text{Tr}}(3072,12288,3072,38,n_{\text{ctx}})
$ \\[2pt]
\textit{Text / embedding:}\\
$\displaystyle 
\text{GFLOPs}_{\text{Tr}}(768,3072,768,12,77)
+
\text{GFLOPs}_{\text{Tr}}(4096,10240,4096,24,512)
+ 24\!\cdot\!10240\!\cdot\!4097/10^9
$ \\[2pt]
\textit{Decoder:}
$\displaystyle \text{GFLOPs}_{\text{Dec}}(H, W)$ \\
\vspace{2pt}\end{minipage}
\\
\midrule
\textbf{Qwen (image)} &
$\displaystyle n_{\text{ctx}}=\frac{HW}{16^2}+12$ &
\begin{minipage}[t]{0.64\linewidth}\small
Image latents ($H/16\times W/16$) plus variable text tokens (averaged to 12 from dataset).  
Hidden width $d_{\text{model}}{=}3072$, $d_{\text{ff}}{=}12288$, $d_{\text{attn}}{=}3072$, $n_{\text{layer}}{=}60$.  
Uses Qwen2.5-VL encoder with MQA (multi-query attention) correction.
\\[4pt]
\textit{Core (diffusion):}
$\displaystyle 
\text{GFLOPs}_{\text{MMDiT}}(3072,12288,3072,60,n_{\text{ctx}})
$ \\[2pt]
\textit{Text / embedding:}
$\displaystyle 
\text{GFLOPs}_{\text{Tr}}(3584,18944,3584,28,12)
+ 28\!\cdot\!12\!\cdot\!2(2\!\cdot\!3584(512-3584))/10^9
$ \\[2pt]
\textit{Decoder:}
$\displaystyle \text{GFLOPs}_{\text{Dec}}(H, W)$ \\
\vspace{2pt}\end{minipage}
\\
\midrule
\textbf{SD\,3.5} &
$\displaystyle n_{\text{ctx}}=\frac{HW}{16^2}+333$ &
\begin{minipage}[t]{0.64\linewidth}\small
Latent grid ($H/16\times W/16$) plus 333 fixed conditioning tokens.  
Hidden width $d_{\text{model}}{=}2432$, $d_{\text{ff}}{=}9478$, $d_{\text{attn}}{=}2432$, $n_{\text{layer}}{=}38$.  
Uses three text encoders: two CLIP variants and T5.
\\[4pt]
\textit{Core (diffusion):}
$\displaystyle 
\text{GFLOPs}_{\text{MMDiT}}(2432,9478,2432,38,n_{\text{ctx}})
$ \\[2pt]
\textit{Text / embedding:}
$\displaystyle 
\text{GFLOPs}_{\text{Tr}}(768,3072,768,12,77)
+
\text{GFLOPs}_{\text{Tr}}(1280,5120,1280,32,77)
+
\text{GFLOPs}_{\text{Tr}}(4096,10240,4096,24,256)
+ 24\!\cdot\!10240\!\cdot\!4097/10^9
$ \\[2pt]
\textit{Decoder:}
$\displaystyle \text{GFLOPs}_{\text{Dec}}(H, W)$ \\
\vspace{2pt}\end{minipage}
\\
\bottomrule
\end{tabular}%
}
\end{table*}




\begin{figure*}[p]
    \centering
    
    \begin{subfigure}[b]{0.08\textwidth}
        \centering
        \rotatebox{90}{\textbf{Training}}
        \vspace{1.5cm}
    \end{subfigure}
    \begin{subfigure}[b]{0.29\textwidth}
        \centering
        \includegraphics[height=3.5cm]{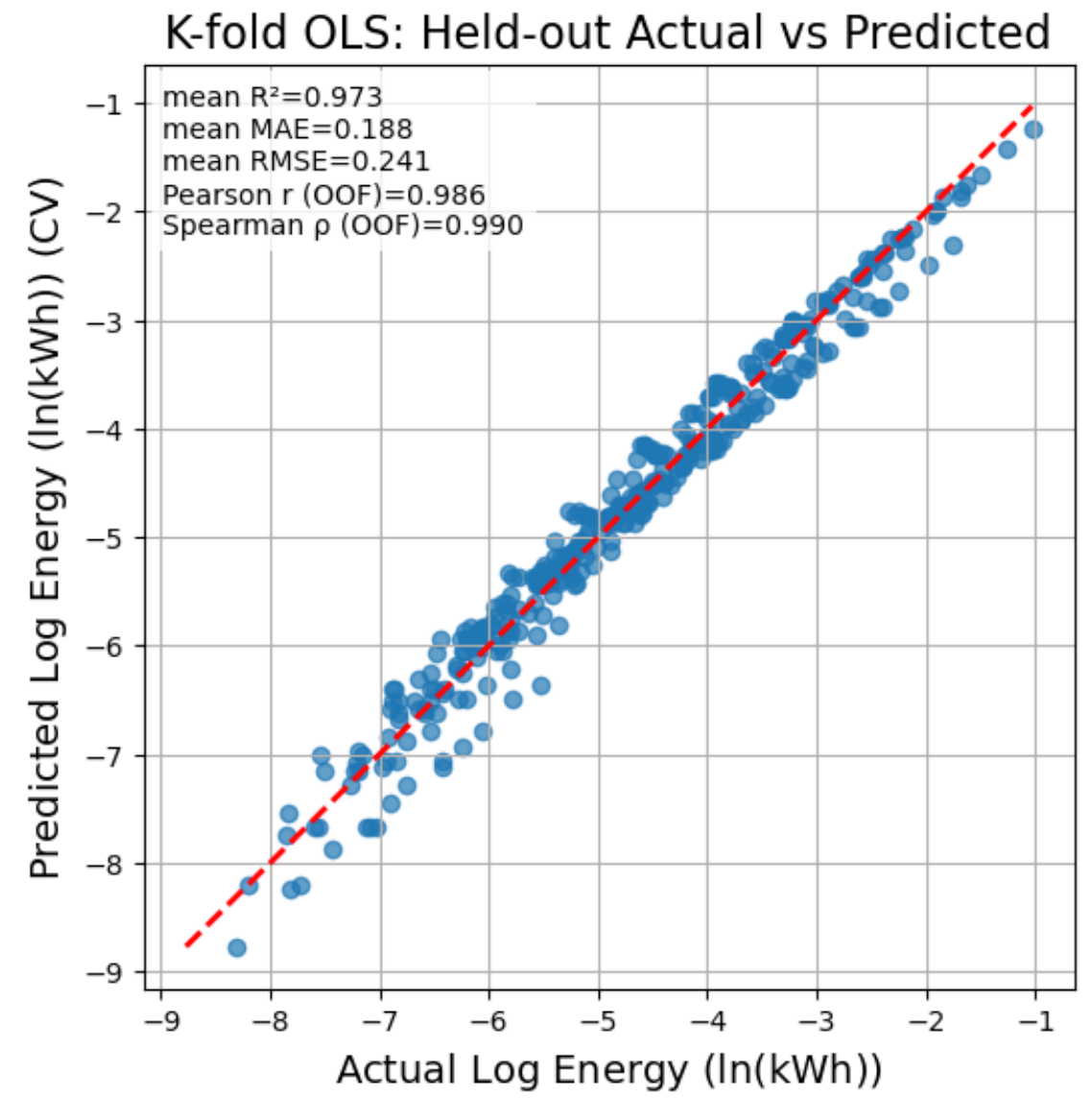}
        \caption{Qwen + SD2}
        \label{fig:qwen_sd2_pair}
    \end{subfigure}
    \hfill
    \begin{subfigure}[b]{0.29\textwidth}
        \centering
        \includegraphics[height=3.5cm]{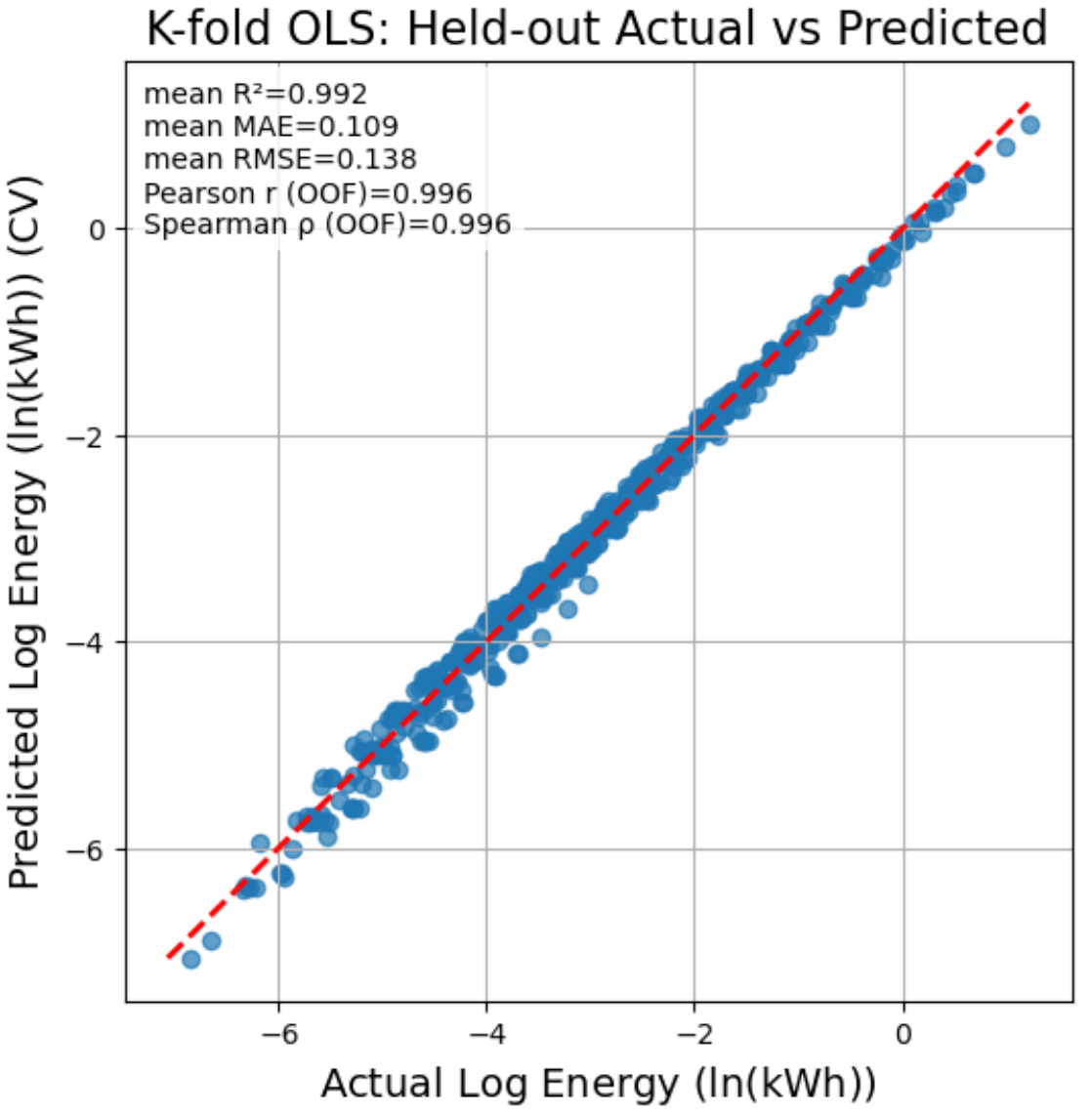}
        \caption{Flux + Qwen + SD3.5}
        \label{fig:flux_qwen_sd35_pair}
    \end{subfigure}
    \hfill
    \begin{subfigure}[b]{0.29\textwidth}
        \centering
        \includegraphics[height=3.5cm]{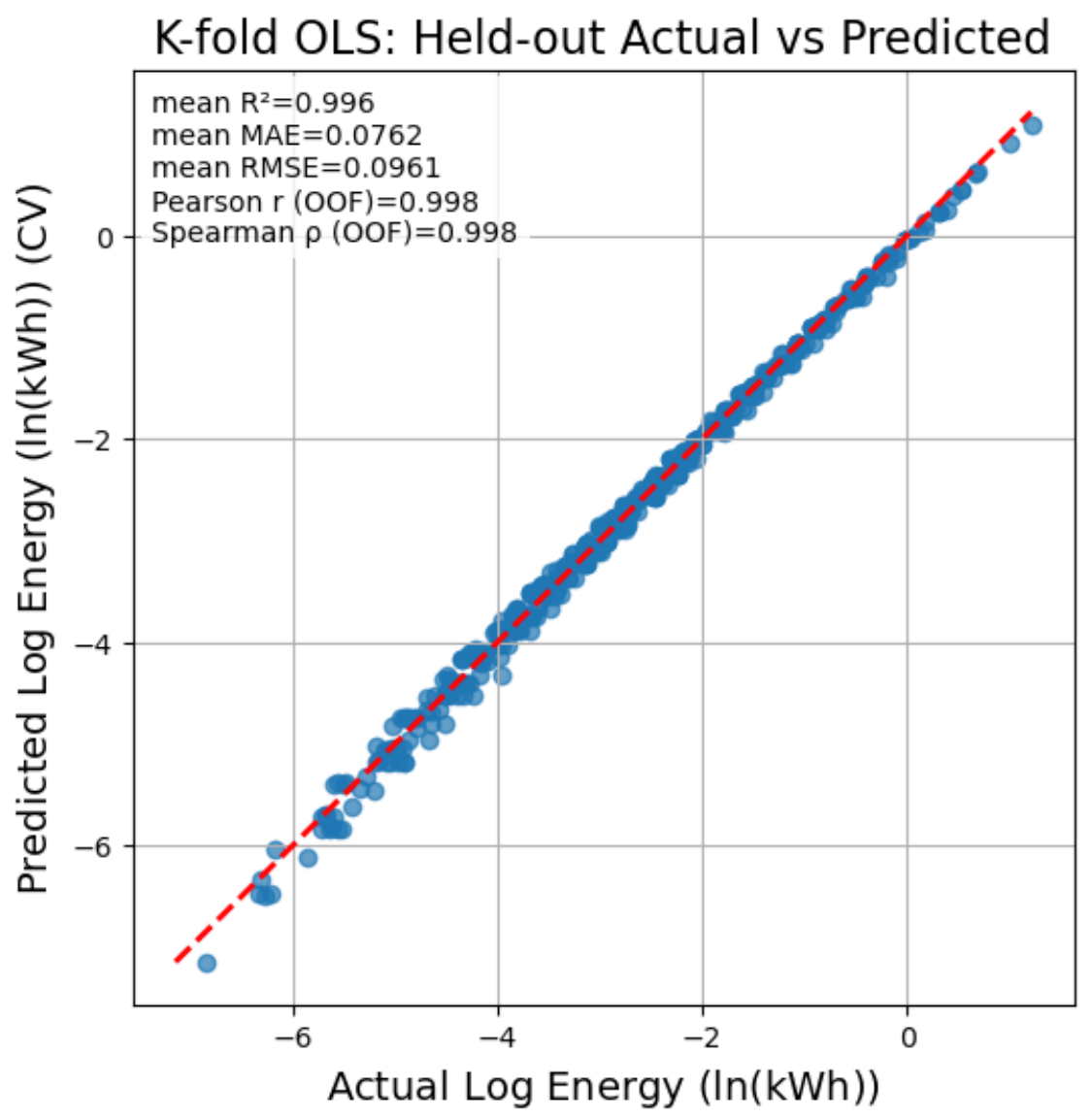}
        \caption{SD3.5 + Flux}
        \label{fig:sd35_flux_pair}
    \end{subfigure}
    
    \vspace{0.2cm}
    
    \begin{subfigure}[b]{0.08\textwidth}
        \centering
        \rotatebox{90}{\textbf{Testing}}
        \vspace{1.5cm}
    \end{subfigure}
    \begin{subfigure}[b]{0.29\textwidth}
        \centering
        \includegraphics[height=3.5cm]{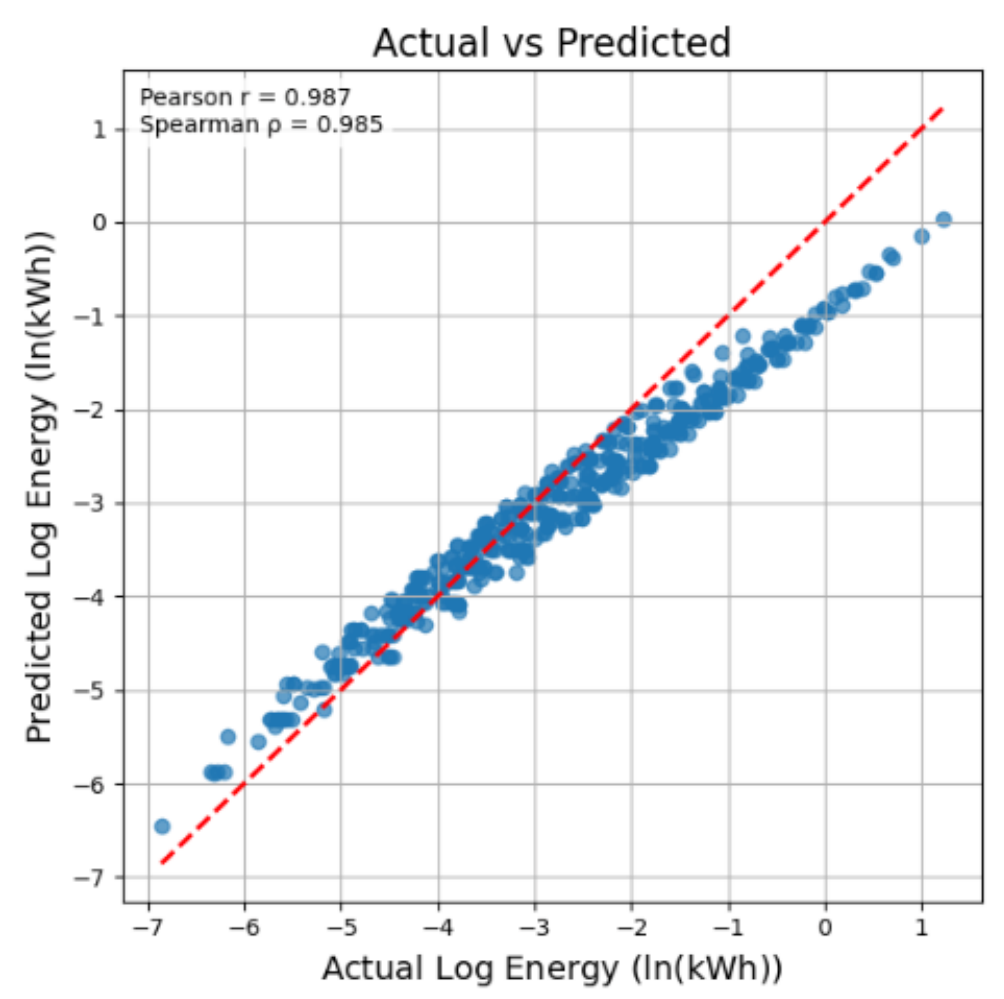}
        \caption*{Flux + SD3.5}
        \label{fig:flux_sd35_test}
    \end{subfigure}
    \hfill
    \begin{subfigure}[b]{0.29\textwidth}
        \centering
        \includegraphics[height=3.5cm]{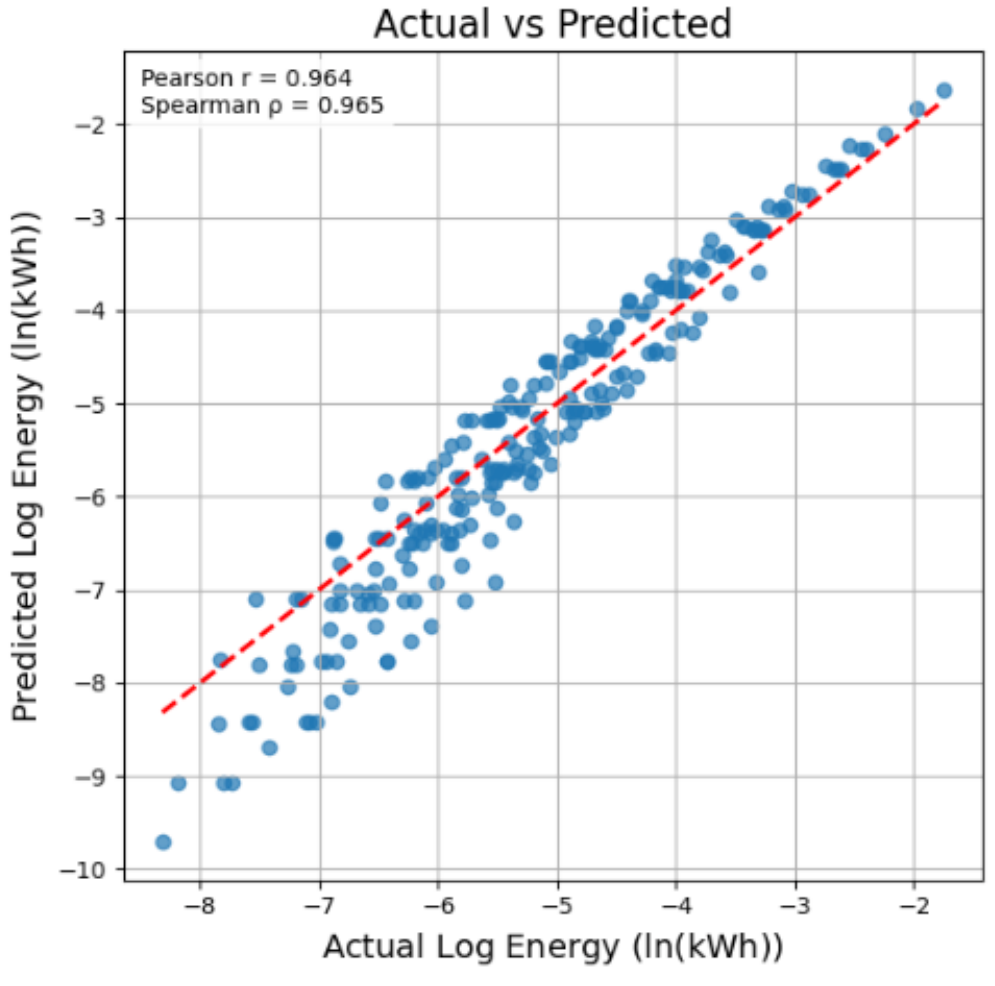}
        \caption*{Stable Diffusion 2}
        \label{fig:sd2_test}
    \end{subfigure}
    \hfill
    \begin{subfigure}[b]{0.29\textwidth}
        \centering
        \includegraphics[height=3.5cm]{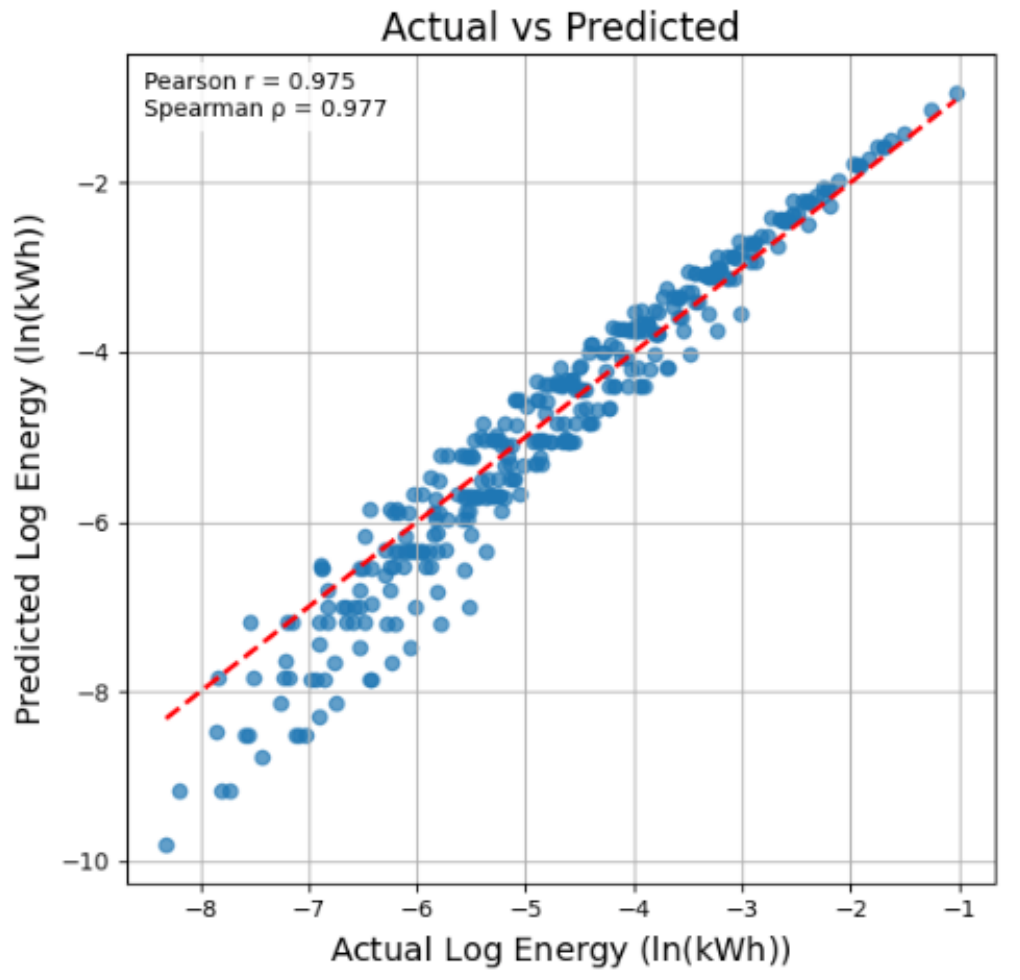}
        \caption*{SD2 + Qwen}
        \label{fig:sd2_qwen_test}
    \end{subfigure}
    
    \caption{Cross-architecture experiments demonstrate generalization between U-Net and MMDiT architectures on the A100. Top row shows training: (a) Qwen+SD2 (MMDiT+U-Net), (b) Flux+Qwen+SD3.5 (all MMDiT), (c) SD3.5+Flux (both MMDiT). Bottom row shows corresponding testing. The consistent scaling patterns validate that our FLOP-based methodology captures fundamental energy-complexity relationships across different architectural paradigms.}
    \label{fig:cross_non_mmdit_architecture_validation}

    \vspace{1.5em} 
    \hrule
    \vspace{1.5em}

    \begin{subfigure}[b]{0.08\textwidth}
        \centering
        \rotatebox{90}{\textbf{Training}}
        \vspace{1.5cm}
    \end{subfigure}
    \begin{subfigure}[b]{0.29\textwidth}
        \centering
        \includegraphics[height=3.5cm]{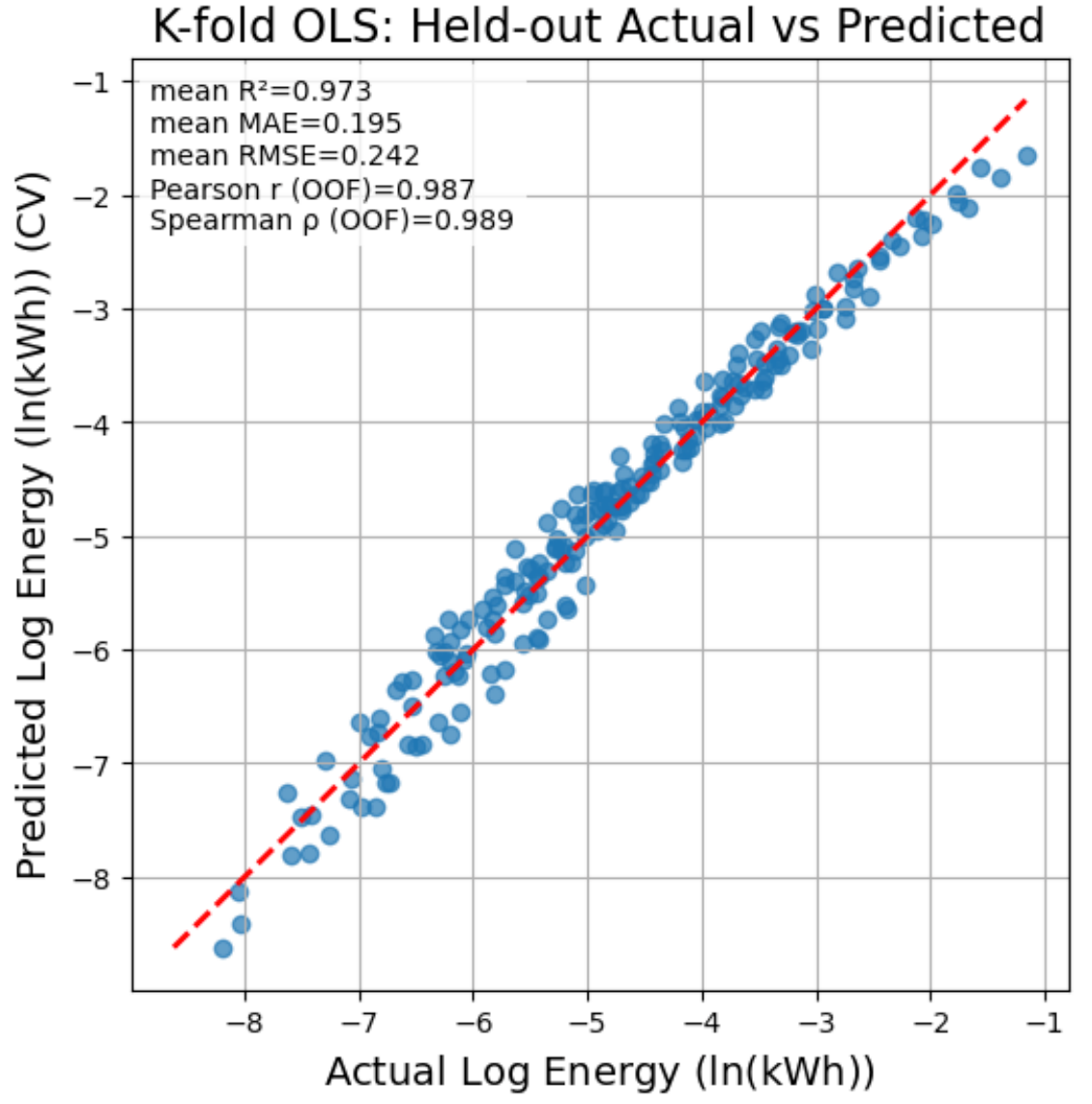}
        \caption{SD2 + SD3.5}
        \label{fig:qwen_sd2_pair_cross}
    \end{subfigure}
    \hfill
    \begin{subfigure}[b]{0.29\textwidth}
        \centering
        \includegraphics[height=3.5cm]{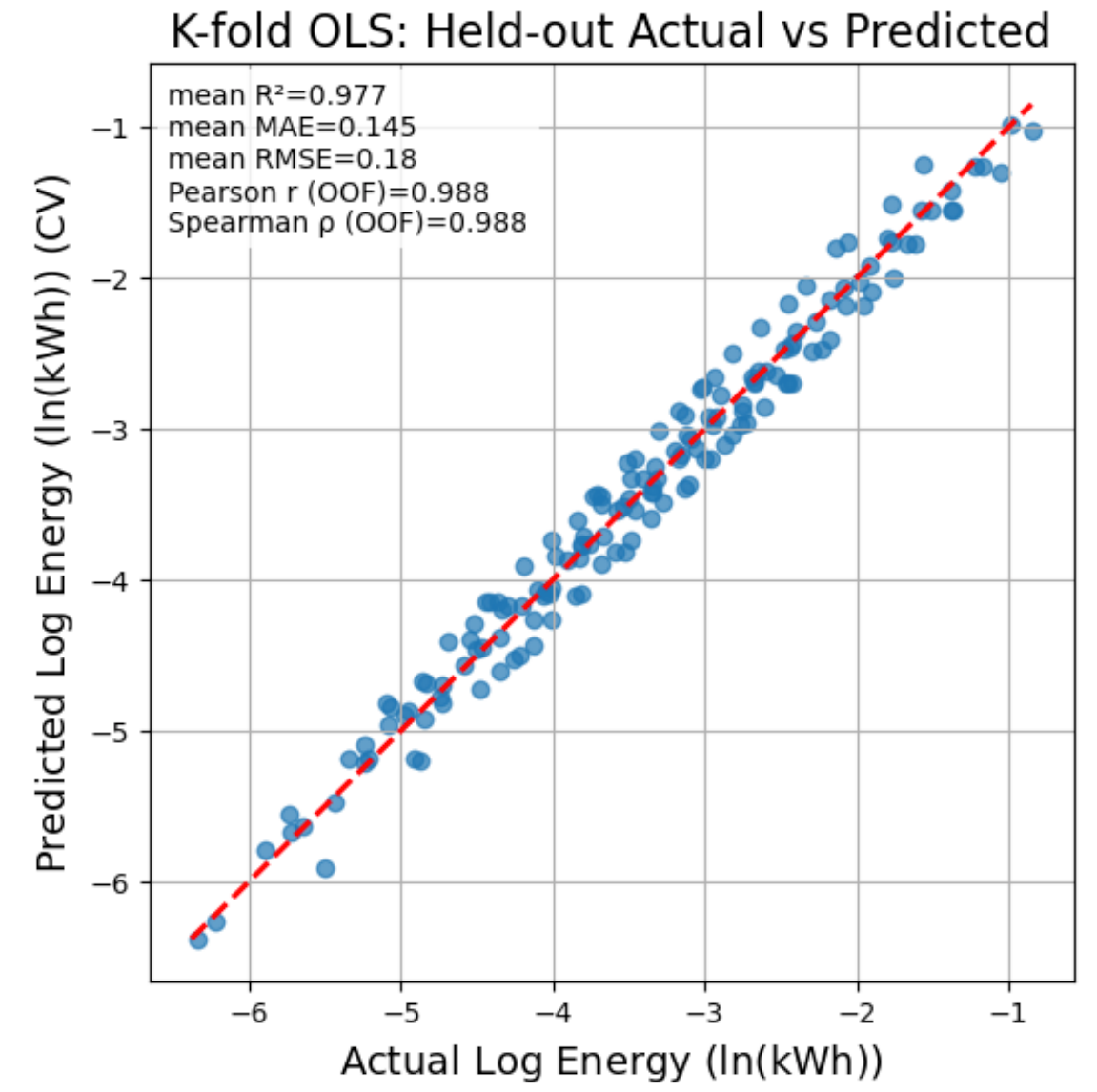}
        \caption{SD3.5 + Flux}
        \label{fig:flux_qwen_sd35_pair_cross}
    \end{subfigure}
    \hfill
    \begin{subfigure}[b]{0.29\textwidth}
        \centering
        \includegraphics[height=3.5cm]{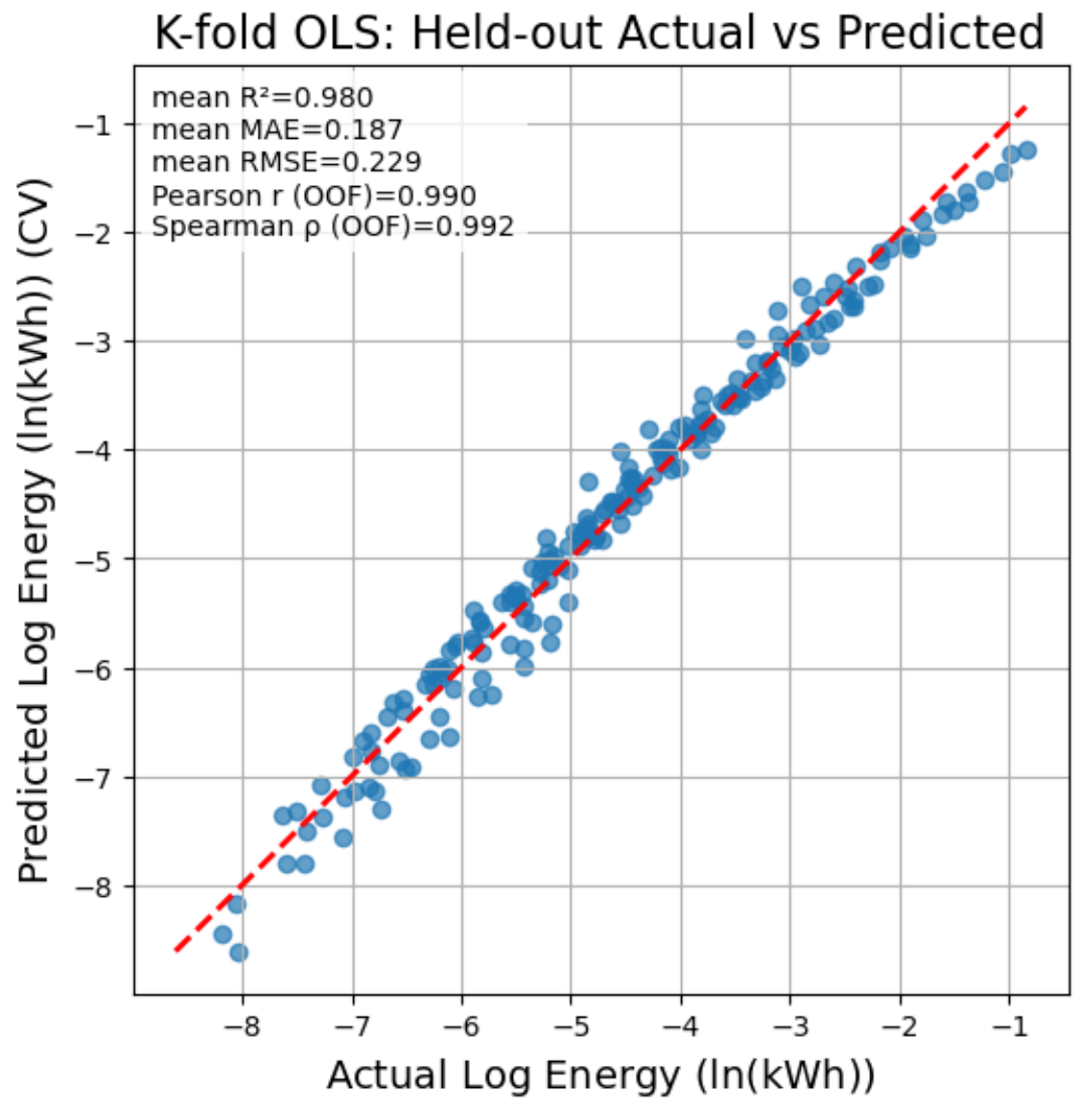}
        \caption{Flux + SD2}
        \label{fig:sd35_flux_pair_cross}
    \end{subfigure}
    
    \vspace{0.2cm}
    
    \begin{subfigure}[b]{0.08\textwidth}
        \centering
        \rotatebox{90}{\textbf{Testing}}
        \vspace{1.5cm}
    \end{subfigure}
    \begin{subfigure}[b]{0.29\textwidth}
        \centering
        \includegraphics[height=3.5cm]{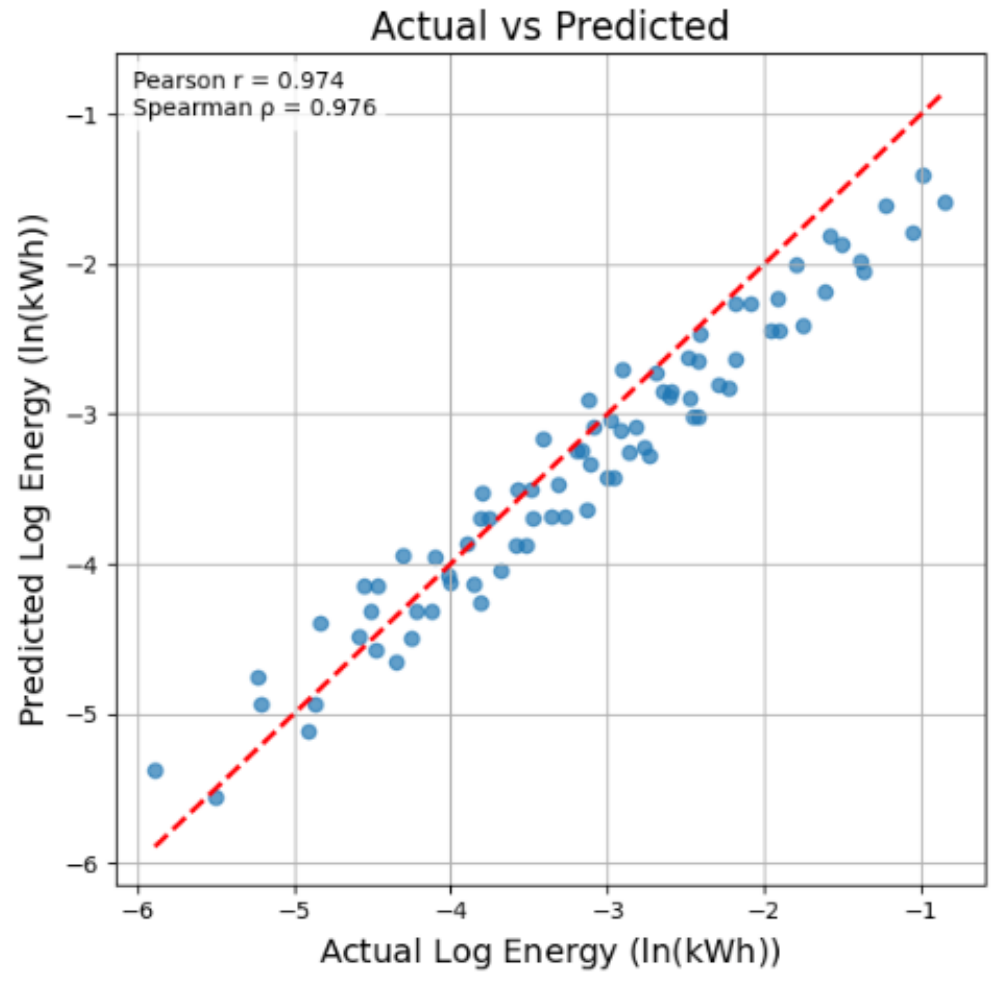}
        \caption*{Flux}
        \label{fig:flux_sd35_test_cross}
    \end{subfigure}
    \hfill
    \begin{subfigure}[b]{0.29\textwidth}
        \centering
        \includegraphics[height=3.5cm]{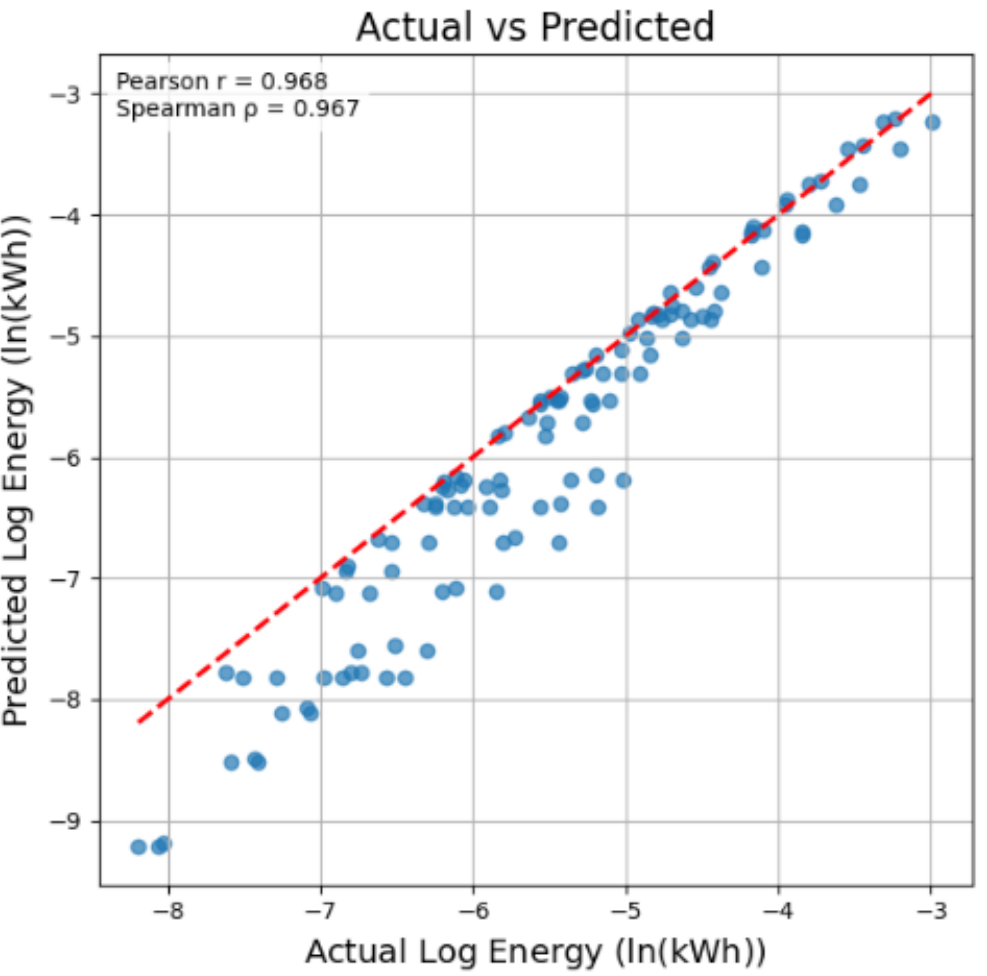}
        \caption*{Stable Diffusion 2}
        \label{fig:sd2_test_cross}
    \end{subfigure}
    \hfill
    \begin{subfigure}[b]{0.29\textwidth}
        \centering
        \includegraphics[height=3.5cm]{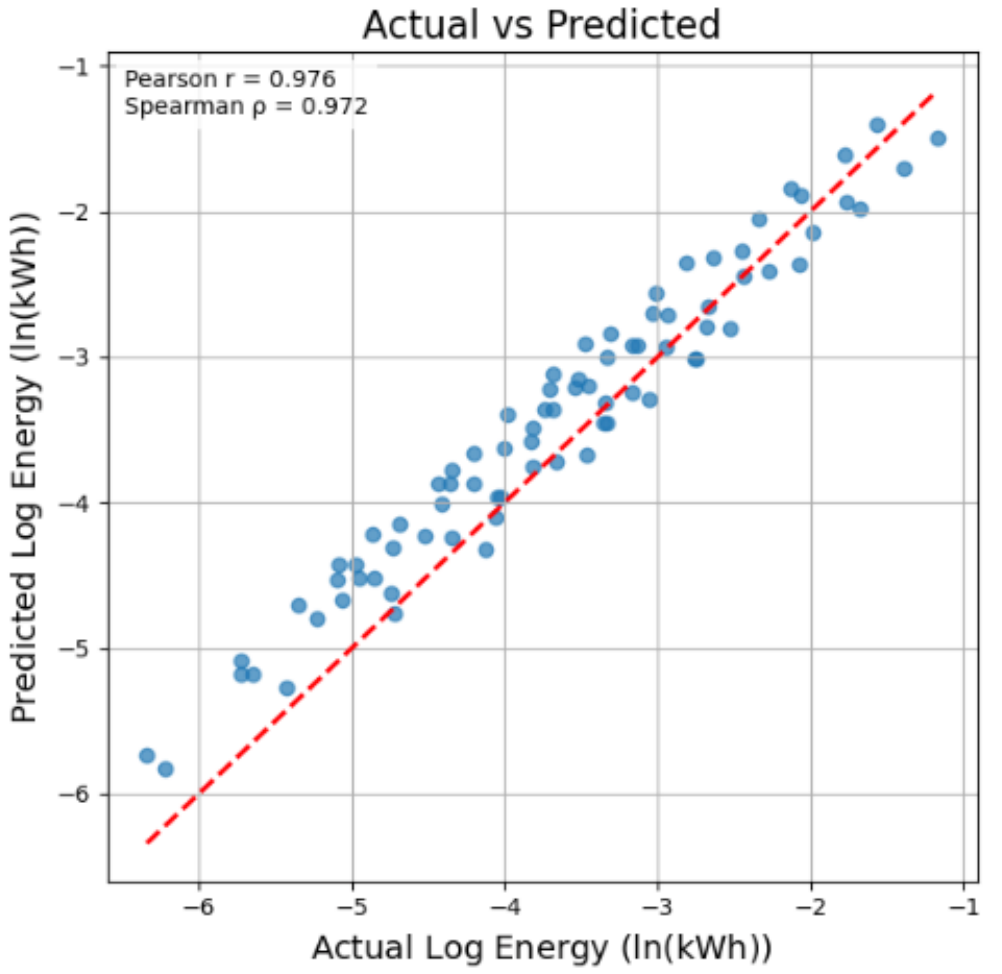}
        \caption*{Stable Diffusion 3.5}
        \label{fig:sd2_qwen_test_cross}
    \end{subfigure}
    
    \caption{Cross-architecture+gpu experiments demonstrate generalization across the A100 and A6000. Top row shows training: (a) SD3.5+SD2, (b) SD3.5+Flux, (c) Flux+SD2. Bottom row shows corresponding testing. The results validate that our methodology captures energy-complexity relationships across both GPU types and architectural designs.}
    \label{fig:cross_non_mmdit_architecture_gpu_validation}
\end{figure*}

\subsection{Detailed Energy Results}
\label{app:appendix_energy}

Tables~\ref{tab:energy_flux}, \ref{tab:energy_sd3_5}, \ref{tab:energy_sd2}, and \ref{tab:energy_qwen} provide raw energy measurements (in Joules) for all evaluated hyperparameter configurations on the NVIDIA A100 platform across 100 prompts.

\begin{table*}[t!]
\centering
\begin{minipage}[t]{0.49\textwidth}
\centering
\renewcommand{\arraystretch}{0.6}
\setlength{\tabcolsep}{4pt}
\scriptsize
\caption{A100 GPU energy (Joules) for hyperparameter settings for FLUX (100 prompts).}
\label{tab:energy_flux}
\begin{tabular}{c|c|cccc}
\toprule
\textbf{Resolution} & \textbf{Steps} &
\textbf{f16 no-CFG} &
\textbf{f16 CFG} &
\textbf{f32 no-CFG} &
\textbf{f32 CFG} \\
\midrule
\multirow{5}{*}{256×256} 
 & 10  & $2.95\times 10^4$ & $5.81\times 10^4$ & $2.13\times 10^5$ & $4.21\times 10^5$ \\
 & 20  & $5.53\times 10^4$ & $1.11\times 10^5$ & $4.17\times 10^5$ & $8.13\times 10^5$ \\
 & 30  & $8.06\times 10^4$ & $1.61\times 10^5$ & $6.19\times 10^5$ & $1.23\times 10^6$ \\
 & 40  & $1.08\times 10^5$ & $2.14\times 10^5$ & $8.15\times 10^5$ & $1.61\times 10^6$ \\
 & 50  & $1.34\times 10^5$ & $2.66\times 10^5$ & $1.02\times 10^6$ & $2.02\times 10^6$ \\
\midrule
\multirow{5}{*}{512×512} 
 & 10  & $5.45\times 10^4$ & $1.06\times 10^5$ & $4.11\times 10^5$ & $8.15\times 10^5$ \\
 & 20  & $1.03\times 10^5$ & $2.05\times 10^5$ & $8.09\times 10^5$ & $1.62\times 10^6$ \\
 & 30  & $1.54\times 10^5$ & $3.04\times 10^5$ & $1.20\times 10^6$ & $2.38\times 10^6$ \\
 & 40  & $2.01\times 10^5$ & $4.06\times 10^5$ & $1.60\times 10^6$ & $3.23\times 10^6$ \\
 & 50  & $2.52\times 10^5$ & $5.07\times 10^5$ & $1.99\times 10^6$ & $4.00\times 10^6$ \\
\midrule
\multirow{5}{*}{768×768} 
 & 10  & $9.38\times 10^4$ & $1.87\times 10^5$ & $7.13\times 10^5$ & $1.42\times 10^6$ \\
 & 20  & $1.85\times 10^5$ & $3.65\times 10^5$ & $1.42\times 10^6$ & $2.83\times 10^6$ \\
 & 30  & $2.75\times 10^5$ & $5.36\times 10^5$ & $2.10\times 10^6$ & $4.27\times 10^6$ \\
 & 40  & $3.62\times 10^5$ & $7.18\times 10^5$ & $2.85\times 10^6$ & $5.67\times 10^6$ \\
 & 50  & $4.46\times 10^5$ & $9.01\times 10^5$ & $3.52\times 10^6$ & $7.00\times 10^6$ \\
\midrule
\multirow{5}{*}{1024×1024} 
 & 10  & $1.61\times 10^5$ & $3.20\times 10^5$ & $1.22\times 10^6$ & $2.42\times 10^6$ \\
 & 20  & $3.17\times 10^5$ & $6.22\times 10^5$ & $2.44\times 10^6$ & $4.88\times 10^6$ \\
 & 30  & $4.72\times 10^5$ & $9.19\times 10^5$ & $3.71\times 10^6$ & $7.23\times 10^6$ \\
 & 40  & $6.25\times 10^5$ & $1.25\times 10^6$ & $4.92\times 10^6$ & $9.75\times 10^6$ \\
 & 50  & $7.65\times 10^5$ & $1.54\times 10^6$ & $6.07\times 10^6$ & $1.21\times 10^7$ \\
\bottomrule
\end{tabular}

\vspace{0.8cm}

\caption{A100 GPU energy (Joules) for different hyperparameter settings for SD3-5 (100 prompts).}
\label{tab:energy_sd3_5}
\begin{tabular}{c|c|cccc}
\toprule
\textbf{Resolution} & \textbf{Steps} &
\textbf{f16 no-CFG} &
\textbf{f16 CFG} &
\textbf{f32 no-CFG} &
\textbf{f32 CFG} \\
\midrule
\multirow{5}{*}{256×256} 
 & 10  & $1.45\times 10^4$ & $2.49\times 10^4$ & $8.09\times 10^4$ & $1.42\times 10^5$ \\
 & 20  & $2.66\times 10^4$ & $4.69\times 10^4$ & $1.56\times 10^5$ & $2.77\times 10^5$ \\
 & 30  & $3.97\times 10^4$ & $6.72\times 10^4$ & $2.36\times 10^5$ & $4.08\times 10^5$ \\
 & 40  & $5.20\times 10^4$ & $9.04\times 10^4$ & $3.08\times 10^5$ & $5.39\times 10^5$ \\
 & 50  & $6.82\times 10^4$ & $1.11\times 10^5$ & $3.82\times 10^5$ & $6.73\times 10^5$ \\
\midrule
\multirow{5}{*}{512×512} 
 & 10  & $2.62\times 10^4$ & $4.60\times 10^4$ & $1.65\times 10^5$ & $3.08\times 10^5$ \\
 & 20  & $5.00\times 10^4$ & $8.88\times 10^4$ & $3.21\times 10^5$ & $6.02\times 10^5$ \\
 & 30  & $7.29\times 10^4$ & $1.32\times 10^5$ & $4.83\times 10^5$ & $9.11\times 10^5$ \\
 & 40  & $9.66\times 10^4$ & $1.77\times 10^5$ & $6.36\times 10^5$ & $1.20\times 10^6$ \\
 & 50  & $1.22\times 10^5$ & $2.16\times 10^5$ & $7.92\times 10^5$ & $1.50\times 10^6$ \\
\midrule
\multirow{5}{*}{768×768} 
 & 10  & $4.78\times 10^4$ & $9.02\times 10^4$ & $3.33\times 10^5$ & $6.02\times 10^5$ \\
 & 20  & $9.21\times 10^4$ & $1.73\times 10^5$ & $6.52\times 10^5$ & $1.21\times 10^6$ \\
 & 30  & $1.39\times 10^5$ & $2.57\times 10^5$ & $9.73\times 10^5$ & $1.81\times 10^6$ \\
 & 40  & $1.80\times 10^5$ & $3.47\times 10^5$ & $1.28\times 10^6$ & $2.38\times 10^6$ \\
 & 50  & $2.28\times 10^5$ & $4.26\times 10^5$ & $1.63\times 10^6$ & $2.97\times 10^6$ \\
\midrule
\multirow{5}{*}{1024×1024} 
 & 10  & $8.16\times 10^4$ & $1.56\times 10^5$ & $5.89\times 10^5$ & $1.08\times 10^6$ \\
 & 20  & $1.57\times 10^5$ & $3.12\times 10^5$ & $1.16\times 10^6$ & $2.17\times 10^6$ \\
 & 30  & $2.32\times 10^5$ & $4.56\times 10^5$ & $1.74\times 10^6$ & $3.24\times 10^6$ \\
 & 40  & $3.09\times 10^5$ & $6.08\times 10^5$ & $2.32\times 10^6$ & $4.30\times 10^6$ \\
 & 50  & $3.87\times 10^5$ & $7.53\times 10^5$ & $2.94\times 10^6$ & $5.33\times 10^6$ \\
\bottomrule
\end{tabular}
\end{minipage}
\hfill
\begin{minipage}[t]{0.49\textwidth}
\centering
\renewcommand{\arraystretch}{0.6}
\setlength{\tabcolsep}{4pt}
\scriptsize
\caption{A100 GPU energy (Joules) for different hyperparameter settings for SD2 (100 prompts).}
\label{tab:energy_sd2}
\begin{tabular}{c|c|cccc}
\toprule
\textbf{Resolution} & \textbf{Steps} &
\textbf{f16 no-CFG} &
\textbf{f16 CFG} &
\textbf{f32 no-CFG} &
\textbf{f32 CFG} \\
\midrule
\multirow{5}{*}{256×256} 
 & 10  & $3.20\times 10^3$ & $3.81\times 10^3$ & $5.84\times 10^3$ & $8.18\times 10^3$ \\
 & 20  & $5.80\times 10^3$ & $7.26\times 10^3$ & $1.05\times 10^4$ & $1.48\times 10^4$ \\
 & 30  & $8.40\times 10^3$ & $1.08\times 10^4$ & $1.61\times 10^4$ & $2.22\times 10^4$ \\
 & 40  & $1.11\times 10^4$ & $1.39\times 10^4$ & $2.05\times 10^4$ & $2.94\times 10^4$ \\
 & 50  & $1.44\times 10^4$ & $1.69\times 10^4$ & $2.70\times 10^4$ & $3.71\times 10^4$ \\
\midrule
\multirow{5}{*}{512×512} 
 & 10  & $5.50\times 10^3$ & $7.88\times 10^3$ & $1.45\times 10^4$ & $2.31\times 10^4$ \\
 & 20  & $1.00\times 10^4$ & $1.44\times 10^4$ & $2.72\times 10^4$ & $4.51\times 10^4$ \\
 & 30  & $1.47\times 10^4$ & $2.08\times 10^4$ & $3.97\times 10^4$ & $6.58\times 10^4$ \\
 & 40  & $1.94\times 10^4$ & $2.79\times 10^4$ & $5.32\times 10^4$ & $8.86\times 10^4$ \\
 & 50  & $2.31\times 10^4$ & $3.50\times 10^4$ & $6.61\times 10^4$ & $1.09\times 10^5$ \\
\midrule
\multirow{5}{*}{768×768} 
 & 10  & $1.07\times 10^4$ & $1.55\times 10^4$ & $3.46\times 10^4$ & $6.02\times 10^4$ \\
 & 20  & $1.90\times 10^4$ & $2.86\times 10^4$ & $6.69\times 10^4$ & $1.17\times 10^5$ \\
 & 30  & $2.69\times 10^4$ & $4.20\times 10^4$ & $9.97\times 10^4$ & $1.74\times 10^5$ \\
 & 40  & $3.56\times 10^4$ & $5.50\times 10^4$ & $1.30\times 10^5$ & $2.32\times 10^5$ \\
 & 50  & $4.36\times 10^4$ & $6.85\times 10^4$ & $1.64\times 10^5$ & $2.85\times 10^5$ \\
\midrule
\multirow{5}{*}{1024×1024} 
 & 10  & $1.99\times 10^4$ & $3.07\times 10^4$ & $7.20\times 10^4$ & $1.35\times 10^5$ \\
 & 20  & $3.38\times 10^4$ & $5.53\times 10^4$ & $1.38\times 10^5$ & $2.56\times 10^5$ \\
 & 30  & $4.71\times 10^4$ & $8.00\times 10^4$ & $2.00\times 10^5$ & $3.80\times 10^5$ \\
 & 40  & $6.19\times 10^4$ & $1.04\times 10^5$ & $2.64\times 10^5$ & $4.99\times 10^5$ \\
 & 50  & $7.61\times 10^4$ & $1.31\times 10^5$ & $3.26\times 10^5$ & $6.26\times 10^5$ \\
\bottomrule
\end{tabular}

\vspace{0.8cm}

\caption{A100 GPU energy (Joules) for different hyperparameter settings for Qwen (100 prompts).}
\label{tab:energy_qwen}
\begin{tabular}{c|c|cc}
\toprule
\textbf{Res.} & \textbf{Steps} & \textbf{f16 no-CFG} & \textbf{f16 CFG} \\
\midrule
\multirow{5}{*}{256×256}
 & 10  & $1.83\times 10^4$ & $3.67\times 10^4$ \\
 & 20  & $3.76\times 10^4$ & $6.96\times 10^4$ \\
 & 30  & $5.25\times 10^4$ & $1.11\times 10^5$ \\
 & 40  & $7.26\times 10^4$ & $1.43\times 10^5$ \\
 & 50  & $8.99\times 10^4$ & $1.75\times 10^5$ \\
\midrule
\multirow{5}{*}{512×512}
 & 10  & $4.24\times 10^4$ & $7.97\times 10^4$ \\
 & 20  & $8.05\times 10^4$ & $1.56\times 10^5$ \\
 & 30  & $1.20\times 10^5$ & $2.48\times 10^5$ \\
 & 40  & $1.66\times 10^5$ & $3.27\times 10^5$ \\
 & 50  & $2.01\times 10^5$ & $4.04\times 10^5$ \\
\midrule
\multirow{5}{*}{768×768}
 & 10  & $7.48\times 10^4$ & $1.43\times 10^5$ \\
 & 20  & $1.46\times 10^5$ & $2.98\times 10^5$ \\
 & 30  & $2.28\times 10^5$ & $4.32\times 10^5$ \\
 & 40  & $2.88\times 10^5$ & $5.72\times 10^5$ \\
 & 50  & $3.53\times 10^5$ & $7.08\times 10^5$ \\
\midrule
\multirow{5}{*}{1024×1024}
 & 10  & $1.36\times 10^5$ & $2.73\times 10^5$ \\
 & 20  & $2.72\times 10^5$ & $5.25\times 10^5$ \\
 & 30  & $3.98\times 10^5$ & $8.02\times 10^5$ \\
 & 40  & $5.36\times 10^5$ & $1.02\times 10^6$ \\
 & 50  & $6.63\times 10^5$ & $1.29\times 10^6$ \\
\bottomrule
\end{tabular}
\end{minipage}
\end{table*}
\end{document}